\documentclass[10pt,twocolumn,letterpaper]{article}

\usepackage[pagenumbers]{cvpr} 

\usepackage{amssymb}
\usepackage{pifont}

\usepackage{symbols}

\usepackage{amsmath,amsfonts,bm}

\def\1{\bm{1}}

\def\sp{space}

\DeclareMathAlphabet{\mathsfit}{\encodingdefault}{\sfdefault}{m}{sl}
\SetMathAlphabet{\mathsfit}{bold}{\encodingdefault}{\sfdefault}{bx}{n}

\newcommand*{\ShowNotes}{} 
\definecolor{darkred}{rgb}{0.7,0.1,0.1}
\definecolor{darkgreen}{rgb}{0.1,0.7,0.1}
\definecolor{cyan}{rgb}{0.7,0.0,0.7}
\definecolor{dblue}{rgb}{0.2,0.2,0.8}
\definecolor{maroon}{rgb}{0.76,.13,.28}
\definecolor{burntorange}{rgb}{0.81,.33,0}
\definecolor{tealblue}{rgb}{0.212,0.459, 0.533}

\definecolor{mypink}{rgb}{0.93359375, 0.62109375, 0.83984375}

\definecolor{pp}{rgb}{0.43921569, 0.18823529, 0.62745098}
\definecolor{rr}{rgb}{0.5254902 , 0.00784314, 0.12941176}
\definecolor{bb}{rgb}{0.09019608, 0.23529412, 0.37647059}
\definecolor{yy}{rgb}{0.49803922, 0.3372549 , 0.0}
\definecolor{gg}{rgb}{0.02352941, 0.3372549 , 0.17647059}

\definecolor{mygray}{rgb}{0.9, 0.9 , 0.9}

\ifdefined\ShowNotes
  \newcommand{\colornote}[3]{{\color{#1}\bf{#2: #3}\normalfont}}
\else
  \newcommand{\colornote}[3]{}
\fi

\newcommand{\eat}[1]{} 

\newcommand{\name}{{\textit{Top2Ground}}\xspace}

\usepackage{booktabs}
\usepackage{makecell}
\usepackage{siunitx} 
\usepackage{nicematrix}
\usepackage{multirow}
\usepackage{booktabs} 
\usepackage[table]{xcolor}
\usepackage{mathtools}
\definecolor{cvprblue}{rgb}{0.21,0.49,0.74}
\usepackage{wrapfig}
\usepackage{verbatim}
\usepackage{listings}
\usepackage{gensymb}
\usepackage[pagebackref,breaklinks,colorlinks,allcolors=cvprblue]{hyperref}
\usepackage{graphicx}
\usepackage{mwe} 
\def\paperID{16537} 
\def\confName{CVPR}
\def\confYear{2026}

\title{\name: A Height-Aware Dual Conditioning Diffusion Model \\ for Robust Aerial-to-Ground View Generation}

\author{
Jae Joong Lee \quad Bedrich Benes \\
Department of Computer Science, Purdue University \\
\{\small\tt lee2161, bbenes\}@purdue.edu
}

\begin{document}

\maketitle

\begin{abstract}
Generating ground-level images from aerial views is a challenging task due to extreme viewpoint disparity, occlusions, and a limited field of view. We introduce \name, a novel diffusion-based method that directly generates photorealistic ground-view images from aerial input images without relying on intermediate representations such as depth maps or 3D voxels. Specifically, we condition the denoising process on a joint representation of VAE-encoded spatial features (derived from aerial RGB images and an estimated height map) and CLIP-based semantic embeddings. This design ensures the generation is both geometrically constrained by the scene's 3D structure and semantically consistent with its content. We evaluate \name on three diverse datasets: CVUSA, CVACT, and the Auto Arborist. 
Our approach shows 7.3\% average improvement in SSIM across three benchmark datasets, showing Top2Ground can robustly handle both wide and narrow fields of view, highlighting its strong generalization capabilities.
\end{abstract}

\section{Introduction}\label{sec:intro}
Aerial imagery captured by UAVs is increasingly used in applications such as urban planning, environmental monitoring, and 3D mapping, due to its scalability and low acquisition cost. With over one million UAVs registered in the U.S.~\cite{faa_drones} and a projected 6.4\% annual growth rate~\cite{us_drone_growth}, the demand for extracting actionable insights from aerial views will likely continue to rise. Yet, many downstream tasks, such as infrastructure inspection, vegetation analysis, or geolocation, require fine-grained ground-level information that is not directly visible from above. Capturing such data in remote, restricted, or hazardous environments remains logistically challenging and costly. It is relatively simple to fly a UAV over a specific area, but retrieving ground-view images is complicated.

This motivates the task of \textit{aerial-to-ground view synthesis}, i.e., generating plausible, photorealistic ground-level images from aerial input images. The viewpoint is learned implicitly from the training data, which contains paired aerial and ground-level images, and our model learns this specific statistical mapping rather than performing a full 3D reconstruction. However, this task is inherently ill-posed due to extreme viewpoint changes, occlusions, and ambiguous ground semantics. Deep neural models can ease this task by learning the correspondence beforehand and estimating the ground views by conditioning on the aerial views. Early methods based on CNNs~\cite{zhai2017predicting, li2021sat2vid} and GANs~\cite{regmi2018cross, tang2019multi, lu2020geometry} learn direct mappings, but often struggle with spatial distortion and semantic inconsistency. More recent approaches introduce intermediate geometric reasoning to improve fidelity, such as density maps~\cite{sat2density} or voxel-based projections~\cite{crossviewdiff}. The main limitation of these methods is that they require additional annotations or introduce significant computational overhead, which limits their scalability and deployment.

We introduce \name, a novel diffusion-based framework that directly synthesizes ground-view images from aerial inputs. Our contribution is a new modeling principle for this task: the fusion of spatial (VAE), geometric (VAE on height map), and semantic (CLIP) information. This approach eliminates the need for complex 3D intermediate representations, such as voxels, a major limitation of prior work. Our approach introduces a \textit{height-aware dual conditioning} that leverages two pre-trained embedding spaces. These modules play distinct and complementary roles: the VAE, using the aerial RGB and height map, acts as an ``architect," capturing fine-grained structural and geometric details. In contrast, CLIP acts as a ``semantic director," understanding the high-level context of the scene. Moreover, we condition the model on both the aerial RGB image and its estimated height map. This height prior acts as a geometric constraint, ensuring the synthesized ground view is geometrically consistent with the specific input aerial image, not just any plausible view. This dual conditioning is a more holistic approach than standard depth control, designed specifically for the extreme viewpoint shift between the aerial and ground domains.

Unlike prior methods, \name is robust to both wide-field (e.g., CVUSA, CVACT) and narrow-field (e.g., AAD) aerial imagery. This generalization capability is essential for real-world UAV scenarios, where camera configurations vary and the field of view may be limited. By operating entirely in the latent space of a pretrained diffusion model and using classifier-free guidance, \name achieves high-quality synthesis while remaining efficient.

We validate \name on three diverse datasets: CVUSA~\cite{cvusa}, CVACT~\cite{cvact}, and the Auto Arborist Dataset (AAD)~\cite{aad}. Our results demonstrate state-of-the-art performance across perceptual, semantic, and pixel-level metrics. Our model achieves an average improvement of 7.3\% in SSIM, and 44.9\% in KID across the three benchmark datasets.

Our contributions are summarized as follows:
\begin{enumerate} 
    \item \textbf{Height-aware diffusion without intermediate representations.} We propose a diffusion-based architecture that synthesizes ground views directly from aerial imagery, eliminating the need for 3D voxel or density map intermediates. 
    \item \textbf{Dual conditioning with semantic and spatial guidance.} Our model jointly leverages CLIP (for semantic context) and VAE (for structural/geometric detail) embeddings extracted from aerial RGB and height maps, enhancing both structural fidelity and semantic alignment. 
    \item \textbf{Robust generalization across wide and narrow aerial views.} \name\ performs consistently across CVUSA, CVACT, and AAD, including challenging non-panoramic, narrow-FOV settings common in UAV-based deployments. 
\end{enumerate}

\section{Related Work}\label{sec:relatedworks}
\noindent\textbf{Diffusion-based image generation}
has become the state-of-the-art in generative modeling by synthesizing high-fidelity images through iterative denoising~\cite{ho2020denoising}. Score-based models~\cite{song2021scorebased}, guided sampling~\cite{dhariwal2021diffusion}, and latent diffusion~\cite{rombach2022high} have improved both sample quality and training efficiency. Beyond unconditional generation, recent approaches incorporate various conditioning mechanisms. Text-guided models such as GLIDE~\cite{nichol2021glide}, Imagen~\cite{saharia2022photorealistic}, and DiffusionCLIP~\cite{kim2022diffusionclip} leverage language embeddings to control semantic outputs. InstructPix2Pix~\cite{instructpix2pix} refines text-to-image alignment through paired image editing. More structured conditioning has been explored via ControlNet~\cite{controlnet}, which adds learnable branches for edge maps and segmentation masks, and Tree-D Fusion~\cite{treedfusion}, which integrates spatial priors for scene-aware control. These advances demonstrate the capacity of diffusion models for flexible and photorealistic image synthesis under various conditioning modalities.

\noindent\textbf{Aerial-to-ground image synthesis.}
Cross-view image synthesis aims to generate ground-level views from aerial inputs, a task complicated by extreme viewpoint changes, occlusions, and limited field-of-view overlap. Early approaches employed CNN-based~\cite{zhai2017predicting, li2021sat2vid} or GAN-based~\cite{regmi2018cross, tang2019multi, lu2020geometry} architectures to learn direct mappings, but they often fail to preserve spatial consistency and fine details. Geometry-guided models have introduced intermediate representations, such as height maps~\cite{shi2022geometry}, density fields~\cite{sat2density}, or voxel reconstructions~\cite{crossviewdiff}, to improve fidelity. However, these methods require accurate geometric priors or computationally expensive estimation, and often struggle to model high-level semantics or generalize across diverse viewpoints. We also note that NeRF-based view synthesis methods are not directly comparable, as they typically require multiple input views, whereas our task is single-image synthesis.

\name departs from prior work by removing the dependency on intermediate geometric representations. Instead, we propose a dual-conditioning latent diffusion framework that jointly uses a CLIP-based semantic encoder and a VAE-based visual encoder, which operates on both aerial RGB images and their estimated height maps. This enables our model to generate ground-level views that are both structurally coherent and semantically aligned, without requiring 3D voxels, density estimation, or depth supervision.

Notably, prior works have focused mainly on panoramic imagery with wide aerial fields of view. To the best of our knowledge, \name\ is the first framework to demonstrate robust generalization across both wide-FOV (CVUSA, CVACT) and narrow-FOV (Auto Arborist Dataset) aerial imagery,  within a unified architecture.

\begin{figure*}[ht]
    \centering
    \includegraphics[width=1.0\linewidth]
    {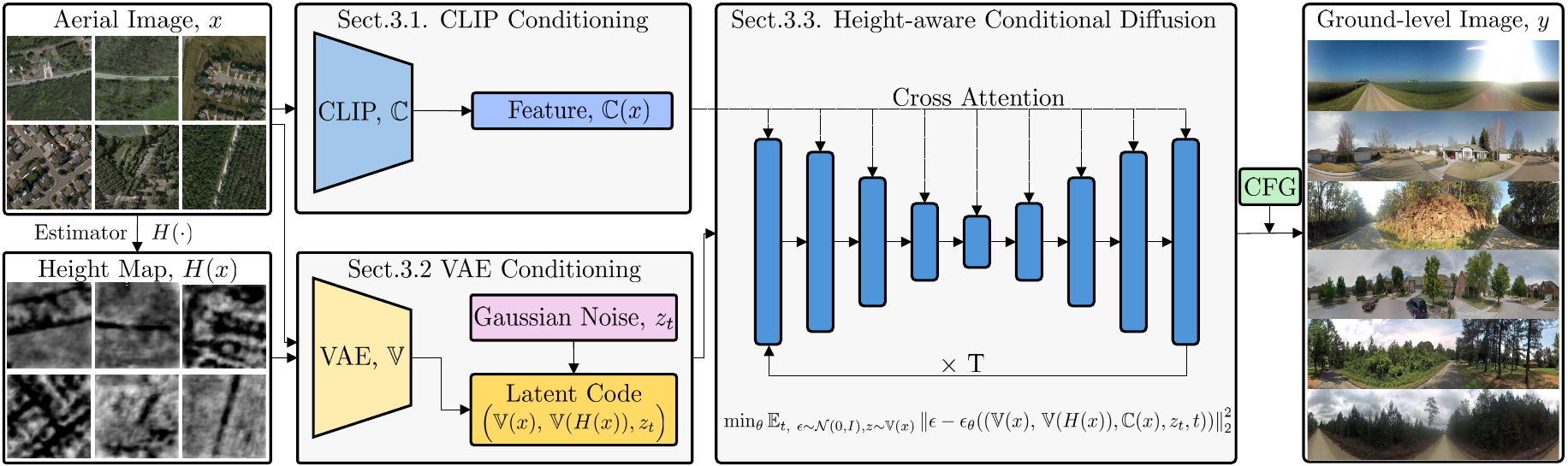}
    \caption{
    \name\ begins by taking an aerial RGB image, $x$, and generating an estimated height map, $H(x)$. $x$ goes into the pre-trained CLIP $\mathbb{C}$ and $x$ and $H(x)$ go into the pre-trained VAE $\mathbb{V}$ to extract semantic and structural embedding features, which are $\mathbb{C}(x)$ and $\mathbb{V}(x)$ respectively. $\mathbb{V}(x)$ is merged with Gaussian Noise $z_t$ to feed in a latent diffusion model, $f_\theta$. In the diffusion process, cross-attention conditioned on $\mathbb{C}(x)$ is utilized to provide semantic consistency. We apply classifier-free guidance with a scale of 2, the model generates a high-quality RGB ground-level image, $y$.
    }
    \label{fig:overview}
    \vspace{-0.5cm}
\end{figure*}

\section{Approach}
\label{sec:approach}
\noindent{\textbf{Task Formulation.}} 
We aim to generate a ground-level view RGB image~$y$ from an aerial RGB image~$x$, which UAVs capture in urban and rural areas. To provide spatial context, we leverage the corresponding estimated height map~$H(x)$ by a pre-trained model~\cite{cambrin2024depth}. The ground-level viewpoint is learned implicitly from the paired aerial and ground-level images in the training dataset. Our model learns this specific statistical mapping and performs a single image-to-image translation, not a 3D reconstruction.

\noindent{\textbf{Overview.}}
We leverage a latent diffusion model~$f_\theta$ conditioned on the image~$x$ and its estimated height map $H(x)$
\begin{equation}
    y = f_\theta(x, H(x)).
\end{equation}
Incorporating $H(x)$ provides spatial context that improves structural details and overall image quality in the generated ground-level view. By fine-tuning a pre-trained Stable Diffusion model~\cite{rombach2022high}, initially trained on Internet-scale data, we focus on learning the domain mapping between aerial and ground-level view imagery, rather than relearning basic image fidelity aspects such as shapes, colors, and textures. This process follows the standard latent diffusion model pipeline. 

(a) During training, the ground-truth ground image $y$ is encoded by the VAE encoder to its latent representation, which the U-Net, $f_\theta$, learns to denoise. The UNet, $f_\theta$, then learns to denoise this latent, conditioned on features from the aerial image and its height map using two pre-trained embedding spaces (VAE~\cite{rombach2022high} and CLIP~\cite{clip}) and classifier-free guidance~\cite{cfg}.
(b) At inference, a Gaussian noise input is iteratively denoised by the model under these same conditions, and the denoised latent is passed through the VAE decoder once to generate the final image $y$.

\subsection{VAE-Based Conditioning}
We extract low-level features $\mathbb{V}(x)$ and $\mathbb{V}(H(x))$ from the aerial RGB image $x$ using the same pre-trained Variational Autoencoder (VAE)~\cite{rombach2022high}. These features capture fine-grained visual details such as texture, color, and spatial structure, which are essential for preserving the visual fidelity of the generated ground-level view images.

We then obtain the noisy latent $z_t$, representing the target image at diffusion time step $t$, and perform a channel-wise concatenation with $\mathbb{V}(x)$.
The channel-wise concatenation of these features with $z_t$ ensures that the diffusion model receives robust low-level details throughout denoising, which is critical for high-ground-level view image fidelity.

\subsection{CLIP-Based Conditioning}
We feed $x$ into a pre-trained CLIP model to extract its semantic features, denoted by $\mathbb{C}(x)$. While the VAE-based conditioning preserves the scene's fine details, the CLIP-based features provide a semantic abstraction of the aerial image, capturing land patterns, roads, buildings, and vegetation to guide the generative process toward semantically plausible and consistent ground-level views. 

These semantic features are injected into the latent diffusion model $f_\theta$ via cross-attention layers, ensuring high-level contexts in the denoising process. Incorporating these features helps the model generate a ground-level view image~$y$ that maintains high image fidelity and aligns well with the semantic context of the aerial scene.

\subsection{Height-aware Conditional Diffusion}
We obtain a relative height map $H(x)$ from~$x$ using a state-of-the-art height estimator~\cite{RegeCambrin2025}, trained on 62 million aerial images. The model achieves a 41.7\% lower MAE and 42.3\% higher IoU than prior methods on the EarthView dataset~\cite{velazquez2025earthview},  
ensuring high-quality and spatially reliable height maps for downstream conditioning. This height prior is a crucial geometric constraint, ensuring the synthesized ground view is geometrically consistent with the specific input aerial image, not just any plausible view. Next, we extract conditioning information via a pre-trained VAE and a CLIP encoder. The VAE captures fine-grained visual details from the aerial image and its height map, while the CLIP encoder provides high-level semantic information from $x$.

We define our conditioning vector $\mathcal{P}$ as
\begin{equation}
{\mathcal{P}} = \left( \mathbb{V}(x),\, \mathbb{V}(H(x)), \mathbb{C}(x), z_t, t \right)),    
\end{equation}
where $\mathbb{V}(x)$ and $\mathbb{V}(H(x))$ are the VAE embeddings of the aerial RGB image and its height map. To process the height map, the single-channel output is replicated three times to match the VAE's 3-channel input. Although trained on RGB images, the VAE's early layers act as meaningful generic extractors for low-level spatial features like edges and textures, which a height map is rich in. This provides an effective geometric prior, as validated by our ablation study~\cref{tab:rebut_height_ab}.
$\mathbb{C}(x)$ is the CLIP embedding of $x$, and~$z_t$ is the noisy latent representation of the ground-level view image $y$ at time step $t$.
These latent embeddings, $\mathbb{V}(x)$ and $\mathbb{V}(H(x))$, already match the spatial dimensions of the noisy latent~$z_t$, allowing them to be concatenated along the channel axis. No additional downsampling of the latent is performed. This ensures that feature maps are aligned for joint conditioning in the denoising network.

In $\mathcal{P}$, the VAE components act as an ``architect,'' supplying fine-grained structural and geometric details, while the CLIP component acts as a ``semantic director'', injecting high-level semantic context. As confirmed by our ablation study (\cref{tab:ablation_condition}), this combination ensures the output is both structurally accurate and semantically plausible.

The iterative denoising is guided by~$z_t$ and~$t$. Inspired by works on novel view synthesis~\cite{liu2023zero, watson2022novel} and conditional generation~\cite{nichol2021glide, controlnet}, we utilize the pre-trained VAE~\cite{rombach2022high} and CLIP~\cite{clip} embedding spaces to condition our latent diffusion model, $f_\theta$. Accordingly, we optimize our model, $f_\theta$, using the following objective with the L2 loss
\begin{equation}
\min_{\theta}\mathbb{E}_{t,\ \epsilon\sim\mathcal{N}(0,I),z\sim\mathbb{V}(x)}\left\|\epsilon-\epsilon_\theta(\mathcal{P})\right\|_2^2,
\end{equation}
where t $\in$ [0, 1,000] denotes the diffusion time step, $\epsilon$ is the Gaussian noise, $\epsilon_\theta$ is the noise prediction network. The model generates $y$ at inference using iterative denoising with a Gaussian noise input~$z_t$ using the conditioning vector~$\mathcal{P}$. 

\subsection{Classifier Free Guidance}
During the training, we randomly nullify the conditional input using classifier-free guidance (CFG)~\cite{cfg}. Our model learns to predict the noise for both conditioned and unconditioned cases. This dual learning enables using a guidance scale during inference to control the influence of the conditioning inputs, thus balancing image fidelity with semantic consistency. 

At inference time, the conditional guidance is scaled by a user-defined factor (set to 2) that controls the conditioning. This method has been empirically shown to improve both the quality and the controllability of the generated outputs~\cite{liu2023zero, nichol2021glide, instructpix2pix, poole2022dreamfusion} in diffusion-based methods.

\section{Experiments} 
\label{sec:experiment} 
Our approach leverages dual conditioning inputs within a diffusion-based model. We train our model using PyTorch 2 on an NVIDIA A100 80GB GPU with a batch size of 192, ensuring maximal VRAM utilization for stable training~\cite{ho2020denoising}, using a learning rate of $10^{-4}$ with AdamW~\cite{loshchilov2017decoupled} with 100 epochs. We fine-tune the entire U-Net of the pre-trained Stable Diffusion model, and LoRA is not used. The VAE and CLIP components are kept frozen during training to leverage their powerful pre-trained feature spaces. During inference, our model runs in 0.93 seconds per image on an NVIDIA RTX 4090 (0.03s for height estimation and 0.9s for generation).

\noindent\textbf{Datasets.} \label{sec:dataset} Following the experiment setup of~\citet{sat2density}, we evaluate our model on diverse datasets covering different image styles: CVUSA~\cite{cvusa}, CVACT~\cite{cvact}, and Google Street View images from the Auto Arborist Dataset (AAD)~\cite{aad}. This multi-dataset evaluation demonstrates the robustness of our method across varying image styles. Comprehensive quantitative and qualitative analyses, employing eight distinct metrics, are presented to validate the quality of the generated images. 

\noindent\ding{192} CVUSA~\cite{cvusa} is a widely used large-scale benchmark consisting of 35,532 training pairs and 8,884 testing pairs, each pair comprising a ground-level image and its corresponding aerial view covering varied rural and urban regions in the United States.

\noindent\ding{193} CVACT (Aligned)~\cite{cvact} is based on CVACT~\cite{cvact_original} and it provides correctly aligned image pairs between satellite and ground-view images, including 26,519 training data and 6,288 testing data from both rural and urban areas in Canberra, Australia.

\noindent\ding{194} Auto Arborist Dataset (AAD)~\cite{aad}: This dataset offers large-scale satellite and ground-level view images across 23 U.S. cities, emphasizing tree-centric ground-level views across diverse terrains, including rural, urban, and forested regions. We use 270,000 images for training and 30,000 for testing, processing each image by centering the detected tree based on the annotations and removing extremely blurry images using Laplacian-based sharpness measures, following the procedure in~\citet{treedfusion}. Although another dataset~\cite{zhu2021vigor} is available, its limited coverage (four cities) led us to favor AAD for its broader geographical representation.

\noindent{\textbf{Evaluation Metrics.}} We assess the quality of the generated images using eight metrics. We use a (1) structural similarity index measure (SSIM) for evaluating structural similarity, and distribution-based metrics such as the (2) inception score (IS) and (3) kernel inception distance (KID) help us compare the statistical properties of generated images with those of real ones. Finally, to capture high-level semantic and perceptual features, we utilize (4) Q-Align~\cite{QAlign}, which is based on a pre-trained large multi-modal model to score image quality, (5) CLIP-based similarity, and (6) Learned Perceptual Image Patch Similarity (LPIPS). Using this set of evaluation metrics, we validate generated images in both low- and high-level image quality.

\noindent{\textbf{Baselines.}} We use Sat2Density~\cite{sat2density} and CrossViewDiff~\cite{crossviewdiff} as the state-of-the-art for specifically tailored the aerial-to-ground image synthesis task. We also consider three Diffusion-based image translation models, such as Brownian Bridge Diffusion Models (BBDM)~\cite{BBDM}, ControlNet~\cite{controlnet}, and Instruct Pix2Pix~\cite{instructpix2pix}. We train all the models with the same dataset as we train our model with their best settings for fair comparisons:\newline
\noindent{\ding{182}} BBDM~\cite{BBDM}: A standard diffusion-based image translation model does not enforce that the diffusion process starts from an input image and ends at its target image. However, it uses Brownian Bridge formulation in the diffusion process to apply an explicit condition to ``bridging'' a domain difference from the input and target image.

\noindent{\ding{183}} ControlNet~\cite{controlnet}: A pre-trained diffusion model is capable of generating high image quality, but controlling its generation is challenging. At the same time, it uses an additional conditioning branch to guide its output. We use ControlNet~\cite{controlnet} to generate a ground-level image from an aerial-view image by conditioning a text prompt as \texttt{street view}. We use this generic text-prompted baseline, as standard pre-trained depth ControlNet~\cite{controlnet} is not designed for the extreme viewpoint shift present in this aerial-to-ground task.

\noindent{\ding{184}} Instruct Pix2Pix~\cite{instructpix2pix}: Using a pre-trained stable diffusion model, it allows another textual information to guide the model to generate an image. We use textual conditioning information as \texttt{street view} to generate a ground-level image from its corresponding aerial image. 

\noindent{\ding{185}} Sat2Density~\cite{sat2density}: A GAN-based model, but its internal module generates density maps to use as an intermediate representation, which encodes structural information, including object layouts. Using the density maps, a conditional GAN approach synthesizes the ground-level view given an aerial image. We put the Inception Score as one of the metrics since this work is based on GAN. 

\noindent{\ding{186}} CrossViewDiff~\cite{crossviewdiff}: A diffusion-based model tailored for the aerial-to-ground image synthesis task. It mainly performs a pixel-level image translation.

\subsection{Quantitative Analaysis}
We use CVUSA~\cite{cvusa}, CVACT~\cite{cvact}, and AAD~\cite{aad} datasets as we explained in~\Cref{sec:dataset}. Our comparisons to CrossViewDiff~\cite{crossviewdiff} are limited to the metrics reported in their paper, as their code is not public. However, we use the quantitative metrics from their work to strictly follow their experimental settings. To validate our results, a Wilcoxon signed-rank test comparing \name to the SOTA~\cite{sat2density} confirmed that our improvements are statistically significant (p $<$ .001) for key metrics across all three datasets.

\begin{table}[ht]
    \setlength{\tabcolsep}{2pt}
    \centering
    \resizebox{\linewidth}{!}{%
    \begin{NiceTabular}{l|cccccc}
    \specialrule{.15em}{.05em}{.05em}
    Method & SSIM($\uparrow$) & IS($\uparrow$) & KID($\downarrow$) & Q-Align($\uparrow$) & CLIP($\uparrow$) & LPIPS($\downarrow$)\\
    \midrule
    BBDM                & 0.39 & 1.98 & 0.19 & 1.59 & 0.48 & 0.68\\
    ControlNet          & 0.39 & 1.42 & 0.10 & \textbf{2.25} & \textbf{0.77} & 0.64\\
    Instruct Pix2Pix    & 0.24 & 2.14 & 0.47 & 1.71 & 0.57 & 0.67\\
    Sat2Density         & 0.36 & \underline{2.39} & \underline{0.08} & 1.35 & 0.49 & \underline{0.56}\\
    CrossViewDiff       & \underline{0.37} & -    & -    & -    & -    & -\\
    \hline
    Ours                & \textbf{0.50} & \textbf{2.63} & \textbf{0.06} & \underline{2.12} & \underline{0.75} & \textbf{0.55}\\
    \specialrule{.15em}{.05em}{.05em}
    \end{NiceTabular}
    }
    \vspace{-0.1cm}
    \caption{
    Evaluation metrics on the CVUSA~\cite{cvusa}. The best result is in \textbf{bold} and the second in \underline{underlined}.
    }
    \label{tab:metrics_cvusa}
    \vspace{-0.3cm}
\end{table}

\noindent{\textbf{CVUSA}} dataset~\cite{cvusa}. Our model outperforms five competitive baseline methods in six out of eight metrics, highlighting its effectiveness in bridging the aerial-to-ground domain gap (see~\Cref{tab:metrics_cvusa}). Our approach has a 6.4\% improvement in Structural Similarity (SSIM), which shows better preservation of the spatial structure, and a 10\% improvement in the Inception Score (IS), indicative of better image fidelity and diversity. Furthermore, a 25\% reduction in the KID suggests that the distribution of generated images is much closer to that of real images. Also, a 1.8\% gain in LPIPS shows our generated images have a better perceptual quality. These improvements demonstrate that our method preserves fine-grained details and maintains semantic consistency, which results in a more realistic and robust ground-level view synthesis than the baselines.

\begin{table}[ht]
    \setlength{\tabcolsep}{2pt}
    \centering
    \resizebox{\linewidth}{!}{%
    \begin{NiceTabular}{l|cccccc}
    \specialrule{.15em}{.05em}{.05em}
    Method & SSIM($\uparrow$) & IS($\uparrow$) & KID($\downarrow$) & Q-Align($\uparrow$) & CLIP($\uparrow$) & LPIPS($\downarrow$)\\ %
    \midrule
    BBDM                & 0.45 & 1.86 & 0.16 & 1.59 & 0.51 & 0.68\\
    ControlNet          & 0.45 & 1.23 & 0.16 & 2.04 & 0.63 & 0.63\\
    Instruct Pix2Pix    & 0.41 & \underline{2.16} & \underline{0.08} & \underline{2.37} & \underline{0.83} & 0.60\\
    Sat2Density         & \underline{0.48} & 2.00 & 0.15 & 1.16 & 0.43 & \underline{0.56}\\
    CrossViewDiff       & 0.41 & - & - & - & - & -\\
    \hline
    Ours                & \textbf{0.51} & \textbf{2.34} & \textbf{0.01} & \textbf{2.48} & \textbf{0.87} & \textbf{0.53}\\
    \specialrule{.15em}{.05em}{.05em}
    \end{NiceTabular}
    }
    \vspace{-0.1cm}
    \caption{
    Evaluation metrics on the CVACT~\cite{cvact}. The best result is in \textbf{bold} and the second in \underline{underlined}.
    }
    \label{tab:metrics_cvact}
    \vspace{-0.3cm}
\end{table}

\noindent\textbf{CVACT.}
We evaluate \name\ on the CVACT test split across six metrics covering structural fidelity, semantic consistency, and perceptual realism (\Cref{tab:metrics_cvact}). Our model outperforms across all criteria, demonstrating generalization to diverse geographies and ground-view conditions. We observe a 2\% gain in SSIM, indicating improved spatial structure preservation, and an 8.3\% increase in Inception Score, reflecting higher realism and diversity in synthesis. 

Notably, \name reduces KID by 87.5\% compared to prior methods, signifying a much closer match between generated and real image distributions. Semantic alignment is further validated by improvements of 4.6\% in Q-Align~\cite{QAlign} and 4.8\% in CLIP similarity, confirming better global and category-level coherence and lowering LPIPS by 5.7\%, indicating improved perceptual image quality.

\begin{table}[ht]
    \setlength{\tabcolsep}{2pt}
    \centering
    \resizebox{\linewidth}{!}{%
    \begin{NiceTabular}{l|cccccc}
    \specialrule{.15em}{.05em}{.05em}
    Method & SSIM($\uparrow$) & IS($\uparrow$) & KID($\downarrow$) & Q-Align($\uparrow$) & CLIP($\uparrow$) & LPIPS($\downarrow$)\\ %
    \midrule
    BBDM                & \underline{0.35} & 2.61 & 0.13 & 1.77 & 0.28 & 0.69\\
    ControlNet          & 0.20 & 1.43 & 0.22 & \textbf{2.29} & \textbf{0.76} & 0.70\\
    Instruct Pix2Pix    & 0.19 & \textbf{3.53} & \underline{0.09} & 2.01 & \underline{0.61} & 0.71\\
    Sat2Density         & 0.32 & 2.51 & 0.16 & 1.20 & 0.16 & \underline{0.63}\\
    \hline
    Ours                & \textbf{0.37} & \underline{3.00} & \textbf{0.07} & \underline{2.17} & 0.47 & \textbf{0.60}\\
    \specialrule{.15em}{.05em}{.05em}
    \end{NiceTabular}
    }
    \vspace{-0.1cm}
    \caption{
    Evaluation metrics on the Auto Arborist Dataset (AAD)~\cite{aad}. Best result is in \textbf{bold} and the second in \underline{underlined}.
    }
    \label{tab:metrics_aad}
    \vspace{-0.3cm}
\end{table}

\noindent{\textbf{Auto Arborist Dataset.}} We evaluate our approach on 30,000 images from the test split of the Auto Arborist Dataset~\cite{aad}, following the setups from~\citet{treedfusion}, which excludes images that are excessively blurry due to privacy masking. Unlike the settings in CVUSA~\cite{cvusa} and CVACT~\cite{cvact}, the Auto Arborist Dataset presents a more challenging scenario due to its limited field of view. This constraint requires two major points in the generated ground views: high pixel-level accuracy and robust structural consistency. As reported in~\Cref{tab:metrics_aad}, our method achieves a 13.5\% improvement in SSIM. Additionally, our approach delivers a 22.2\% improvement in KID and a 5\% enhancement in LPIPS, indicating its high-quality performance to generate perceptually compelling ground-level views under challenging conditions.

\subsection{Downstream Task Evaluation} \label{sec:downstream} To demonstrate the practical downstream utility of our synthesized images, we tested tree detection using a text-based query object detection model~\cite{liu2024grounding} on the generated images from the AAD test set with the query \texttt{a tree}. Our model's generated images achieved a mean Average Precision (mAP) of 0.72, outperforming the mAP of 0.51 from images generated by the SOTA~\cite{sat2density}. This confirms that the images generated by \name provide more reliable structural and semantic information for downstream applications.

\subsection{Qualitative Analaysis}
\label{sec:qualitative}
We show a side-by-side visual comparison from the input aerial image, ControlNet~\cite{controlnet}, Instruct Pix2Pix~\cite{instructpix2pix} (Inst. Pix2Pix), BBDM~\cite{BBDM}, Sat2Density~\cite{sat2density}, ours and the ground-truth images using CVUSA~\cite{cvusa}, CVACT~\cite{cvact} and Auto Arborist Dataset~\cite{aad}.

\begin{figure*}[ht]
    \centering
    \small
    \setlength{\tabcolsep}{0.5pt} %
    \renewcommand{\arraystretch}{0.1} %
    \begin{tabular}{cccccccc} %
        \multicolumn{1}{c}{Input} &
        \multicolumn{1}{c}{Height Map} &
        \multicolumn{1}{c}{CntrlNet} &
        \multicolumn{1}{c}{Inst P2P} &
        \multicolumn{1}{c}{BBDM} &
        \multicolumn{1}{c}{S2D} &
        \multicolumn{1}{c}{Ours} &
        \multicolumn{1}{c}{GT} \\

        \includegraphics[width=0.125\linewidth]{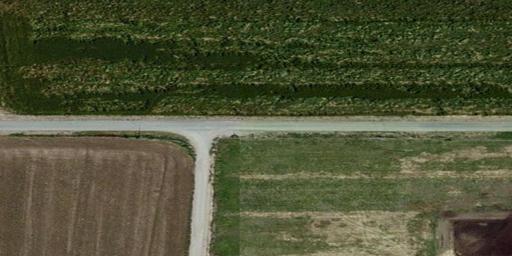} &
        \includegraphics[width=0.125\linewidth]{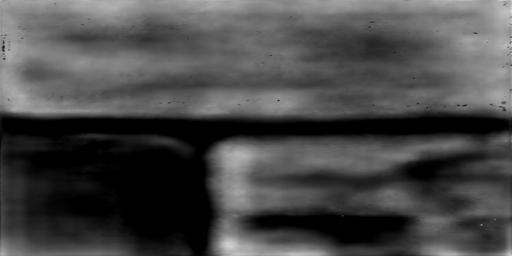} &
        \includegraphics[width=0.125\linewidth]{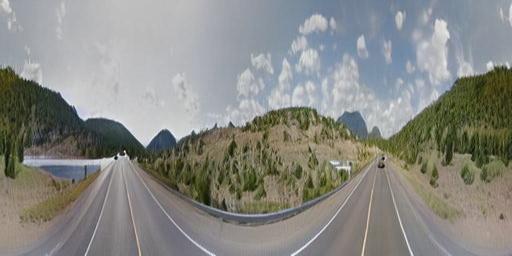} &
        \includegraphics[width=0.125\linewidth]{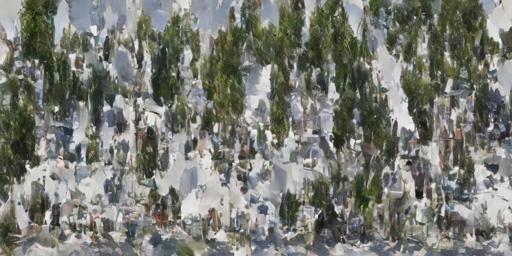} &
        \includegraphics[width=0.125\linewidth]{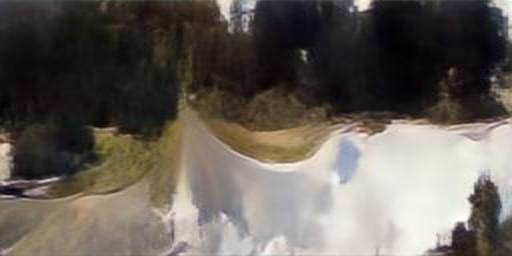} &
        \includegraphics[width=0.125\linewidth]{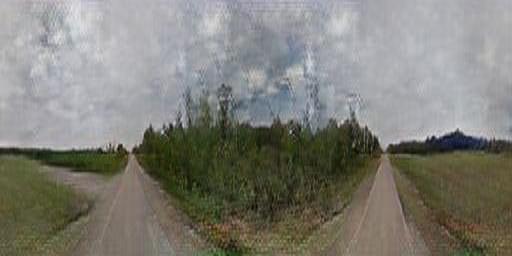} &
        \includegraphics[width=0.125\linewidth]{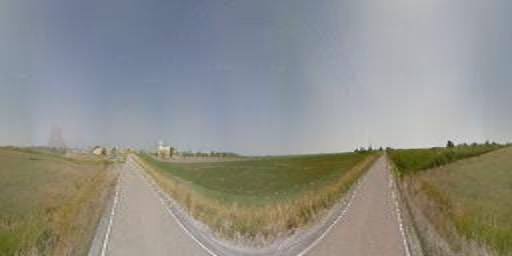} &
        \includegraphics[width=0.125\linewidth]{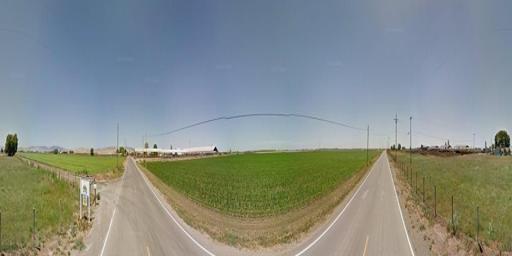}
        \\

        \includegraphics[width=0.125\linewidth]{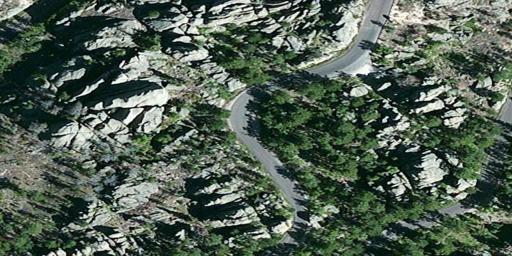} &
        \includegraphics[width=0.125\linewidth]{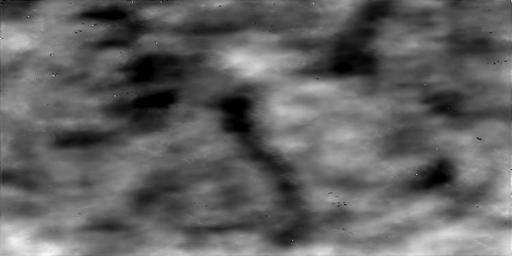} &
        \includegraphics[width=0.125\linewidth]{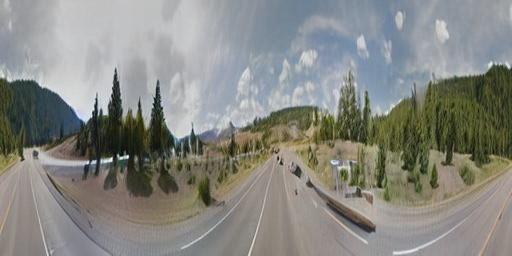} &
        \includegraphics[width=0.125\linewidth]{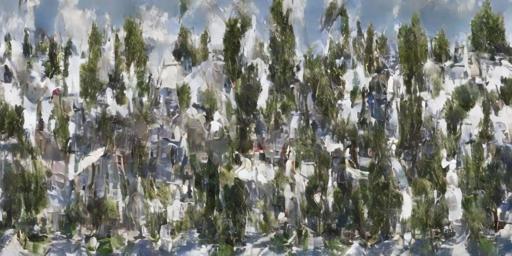} &
        \includegraphics[width=0.125\linewidth]{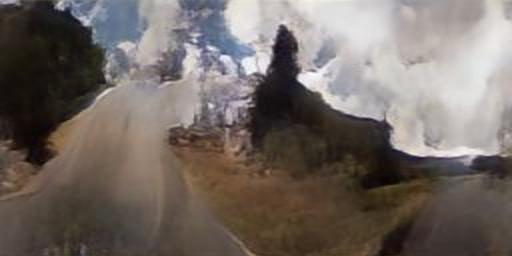} &
        \includegraphics[width=0.125\linewidth]{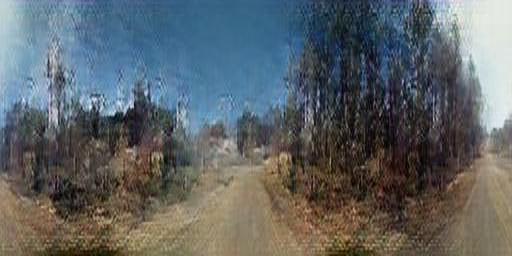} &
        \includegraphics[width=0.125\linewidth]{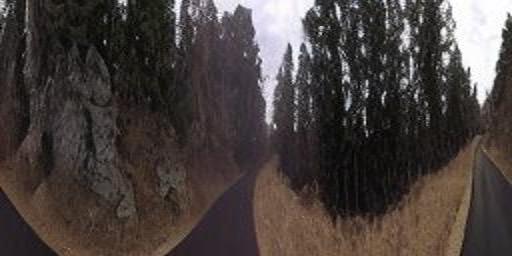} &
        \includegraphics[width=0.125\linewidth]{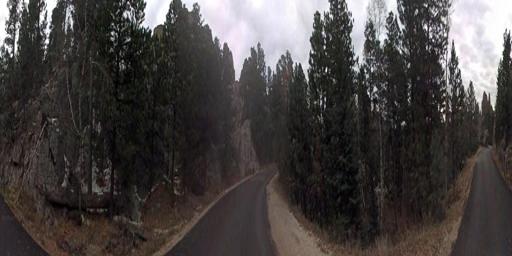}
        \\

        \includegraphics[width=0.125\linewidth]{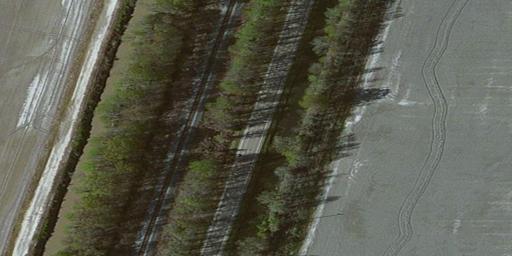} &
        \includegraphics[width=0.125\linewidth]{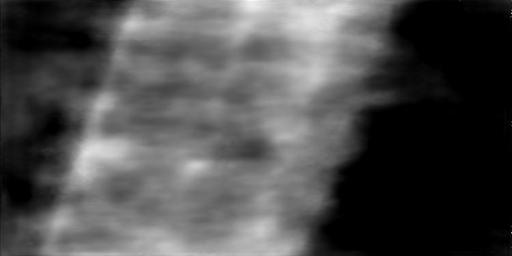} &
        \includegraphics[width=0.125\linewidth]{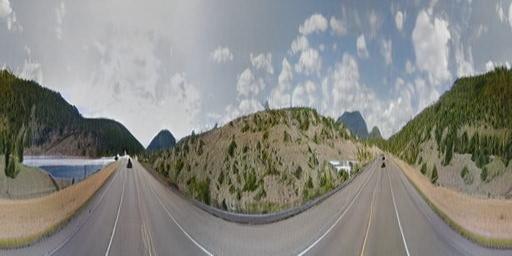} &
        \includegraphics[width=0.125\linewidth]{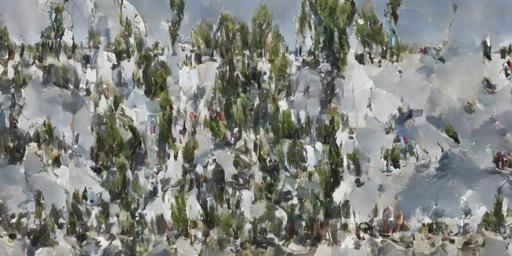} &
        \includegraphics[width=0.125\linewidth]{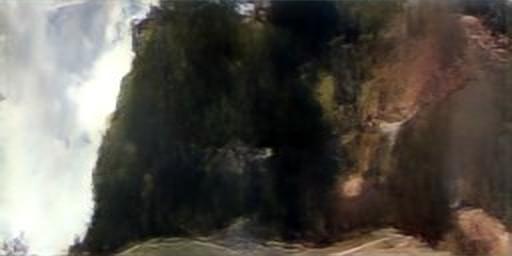} &
        \includegraphics[width=0.125\linewidth]{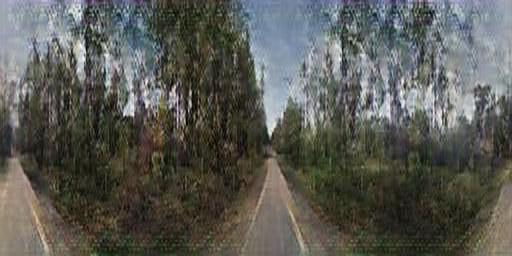} &
        \includegraphics[width=0.125\linewidth]{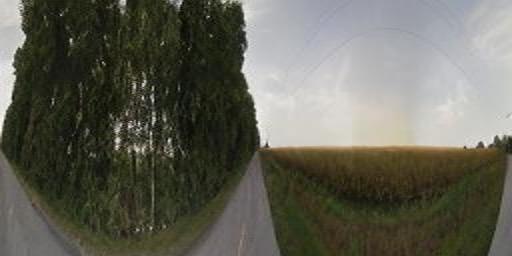} &
        \includegraphics[width=0.125\linewidth]{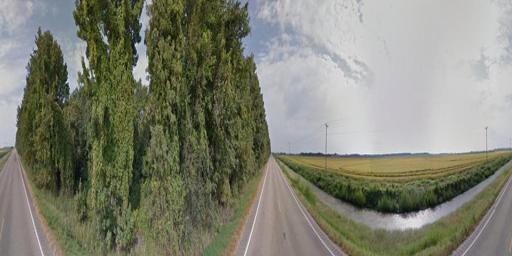}
        \\

        \includegraphics[width=0.125\linewidth]{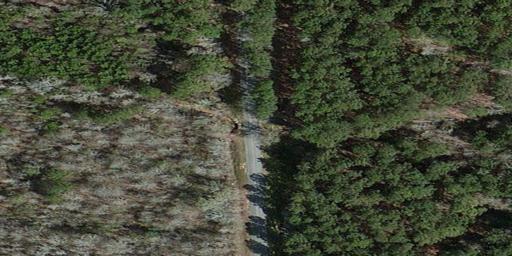} &
        \includegraphics[width=0.125\linewidth]{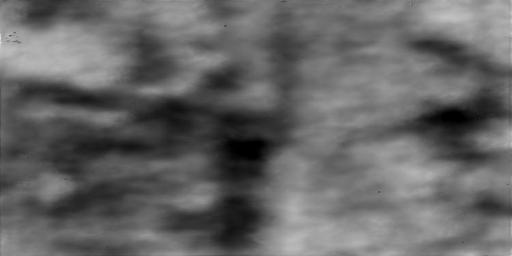} &
        \includegraphics[width=0.125\linewidth]{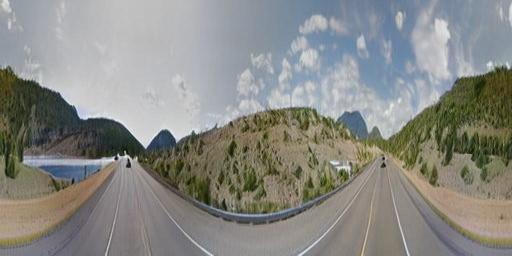} &
        \includegraphics[width=0.125\linewidth]{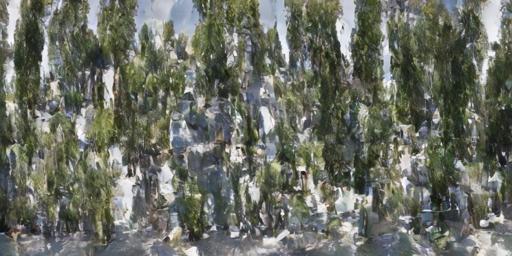} &
        \includegraphics[width=0.125\linewidth]{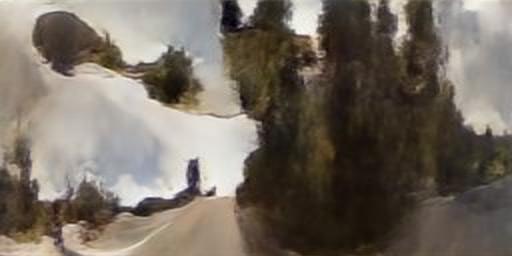} &
        \includegraphics[width=0.125\linewidth]{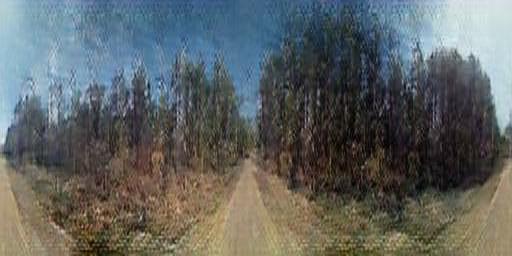} &
        \includegraphics[width=0.125\linewidth]{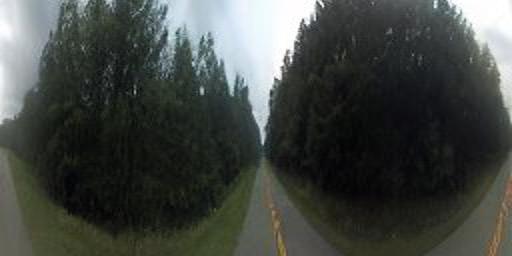} &
        \includegraphics[width=0.125\linewidth]{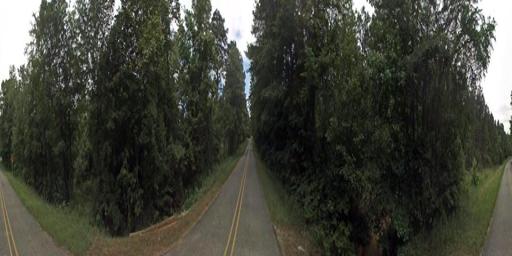}
        \\

    \end{tabular}
        \vspace{-0.2cm}
    \caption{Qualitative comparison of generated ground-level images on the CVUSA dataset. We compare our method with ControlNet (CntrlNet), InstructPix2Pix (Inst P2P), BBDM, and Sat2Density (S2D). Our model better preserves structural layout and semantic coherence, demonstrating improved fidelity and realism over prior approaches.
    }
        \vspace{-0.3cm}
    \label{fig:cvusa_visual}
\end{figure*}

\noindent\textbf{CVUSA.} \figref{fig:cvusa_visual} shows that our height-aware dual conditioning improves structural fidelity in the generated images. All rows demonstrate that leveraging height maps enables our model to understand the scene height accurately. For example, in the third row, our method successfully differentiates between taller trees on the left and shorter trees on the right, whereas Sat2Density~\cite{sat2density} fails to reproduce this distinction, resulting in distorted structural details. 

\begin{figure*}[ht]
    \centering
    \small
    \setlength{\tabcolsep}{0.5pt} %
    \renewcommand{\arraystretch}{0.1} %
    \begin{tabular}{cccccccc} %
        \multicolumn{1}{c}{Input} &
        \multicolumn{1}{c}{Height Map} &
        \multicolumn{1}{c}{CntrlNet} &
        \multicolumn{1}{c}{Inst P2P} &
        \multicolumn{1}{c}{BBDM} &
        \multicolumn{1}{c}{S2D} &
        \multicolumn{1}{c}{Ours} &
        \multicolumn{1}{c}{GT} \\

        \includegraphics[width=0.125\linewidth]{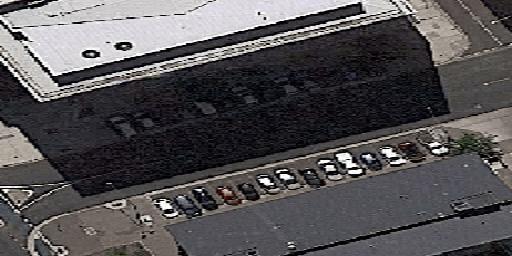} &
        \includegraphics[width=0.125\linewidth]{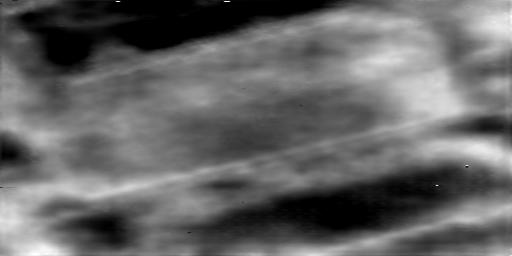} &
        \includegraphics[width=0.125\linewidth]{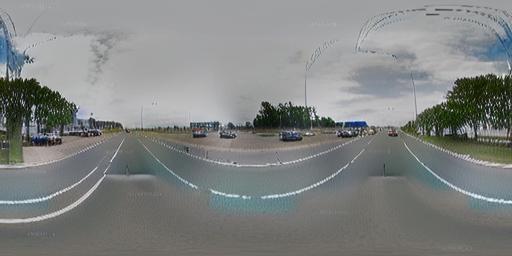} &
        \includegraphics[width=0.125\linewidth]{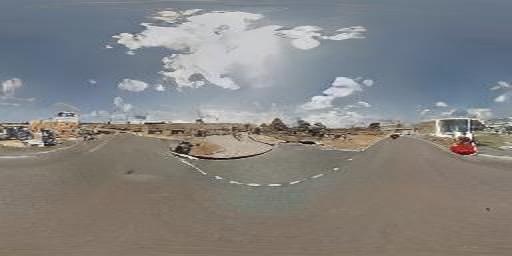} &
        \includegraphics[width=0.125\linewidth]{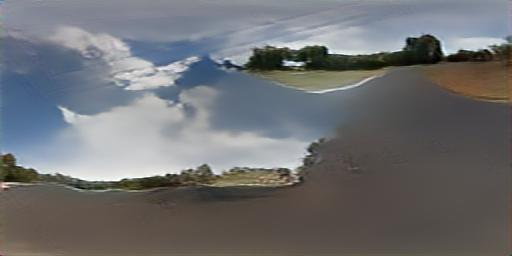} &
        \includegraphics[width=0.125\linewidth]{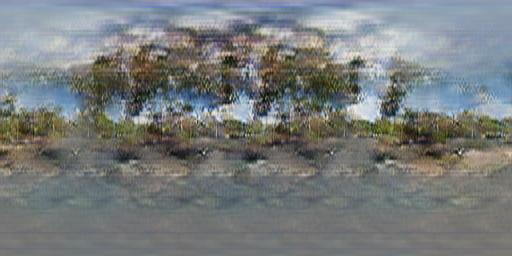} &
        \includegraphics[width=0.125\linewidth]{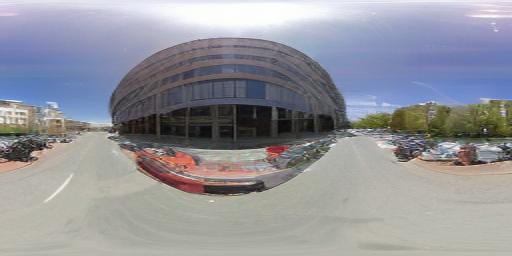} &
        \includegraphics[width=0.125\linewidth]{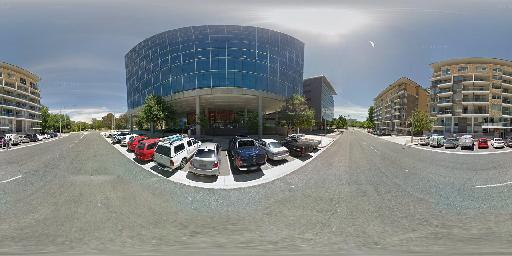}
        \\

        \includegraphics[width=0.125\linewidth]{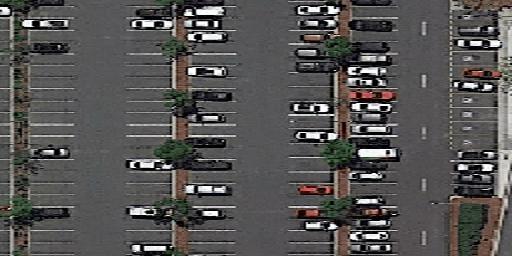} &
        \includegraphics[width=0.125\linewidth]{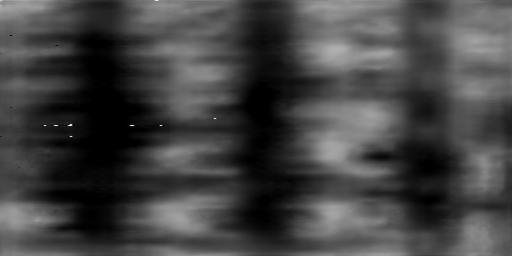} &
        \includegraphics[width=0.125\linewidth]{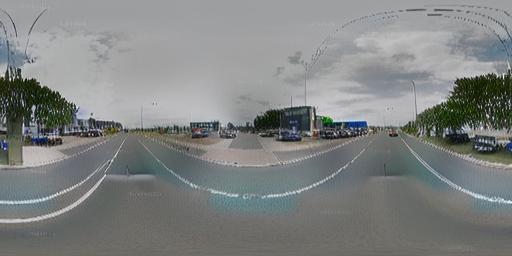} &
        \includegraphics[width=0.125\linewidth]{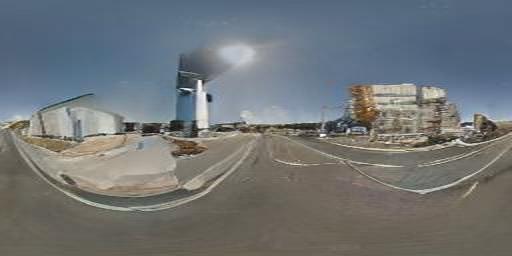} &
        \includegraphics[width=0.125\linewidth]{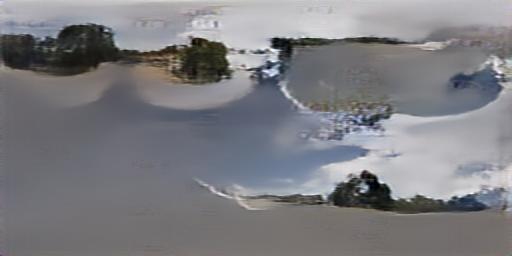} &
        \includegraphics[width=0.125\linewidth]{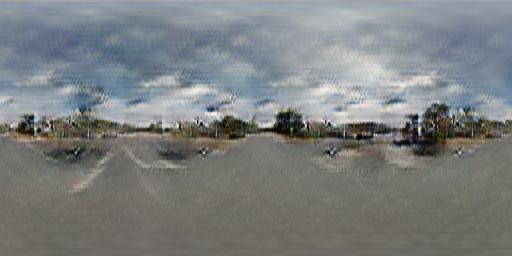} &
        \includegraphics[width=0.125\linewidth]{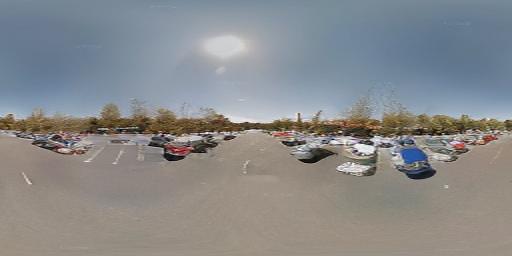} &
        \includegraphics[width=0.125\linewidth]{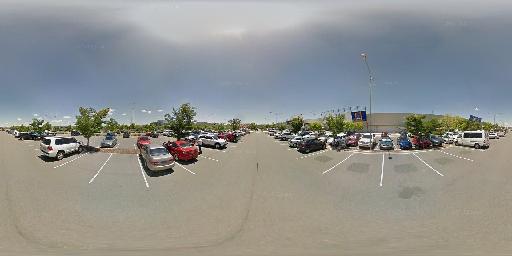}
        \\

        \includegraphics[width=0.125\linewidth]{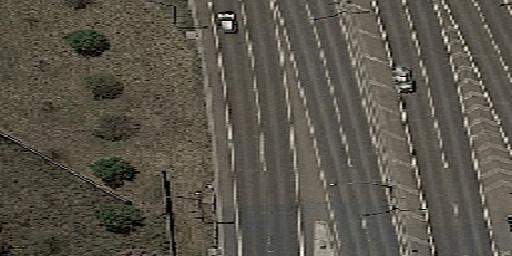} &
        \includegraphics[width=0.125\linewidth]{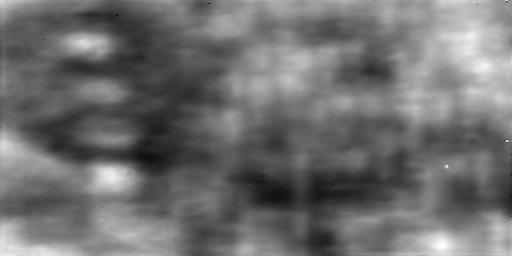} &
        \includegraphics[width=0.125\linewidth]{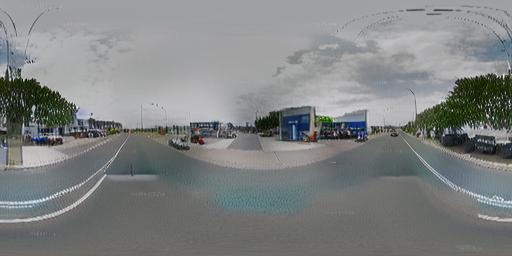} &
        \includegraphics[width=0.125\linewidth]{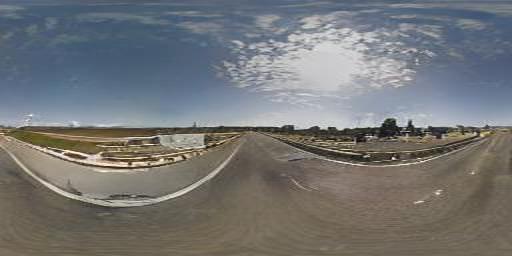} &
        \includegraphics[width=0.125\linewidth]{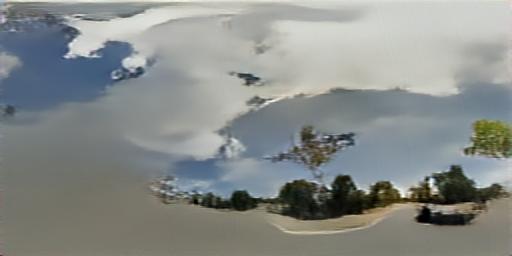} &
        \includegraphics[width=0.125\linewidth]{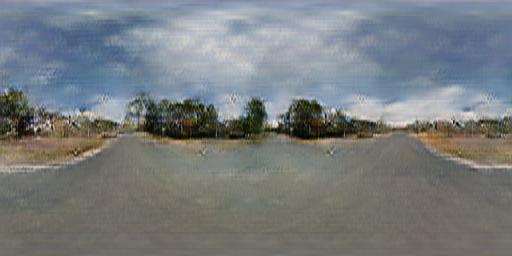} &
        \includegraphics[width=0.125\linewidth]{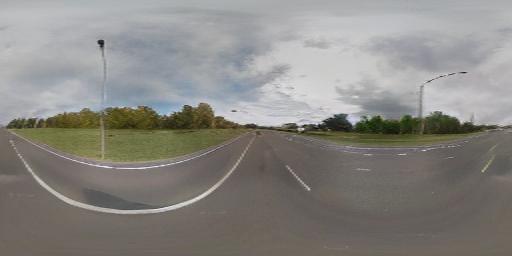} &
        \includegraphics[width=0.125\linewidth]{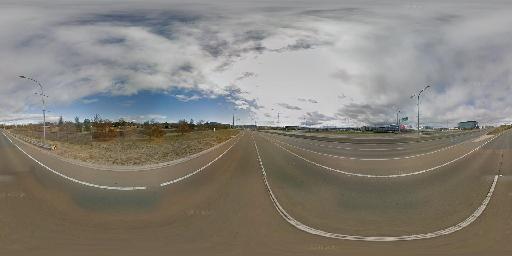}
        \\

        \includegraphics[width=0.125\linewidth]{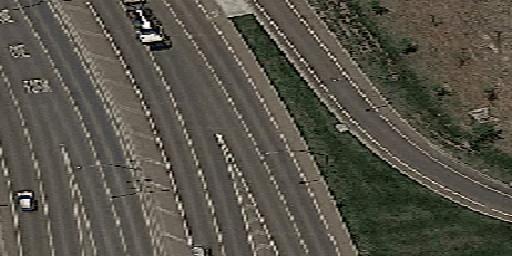} &
        \includegraphics[width=0.125\linewidth]{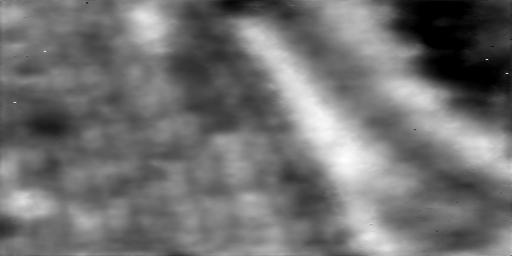} &
        \includegraphics[width=0.125\linewidth]{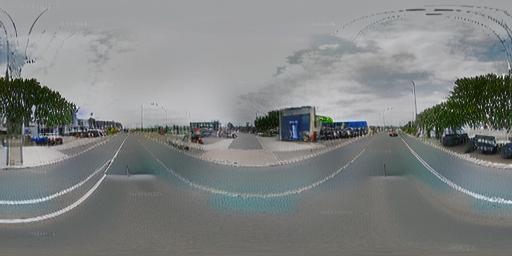} &
        \includegraphics[width=0.125\linewidth]{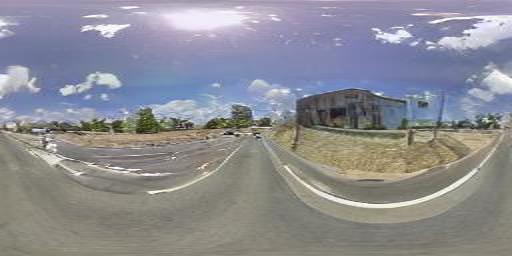} &
        \includegraphics[width=0.125\linewidth]{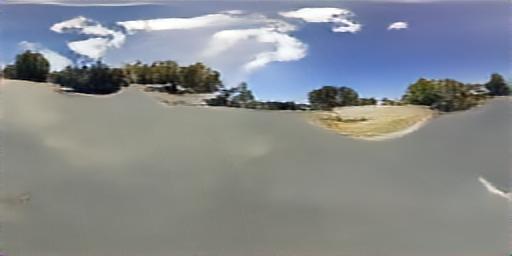} &
        \includegraphics[width=0.125\linewidth]{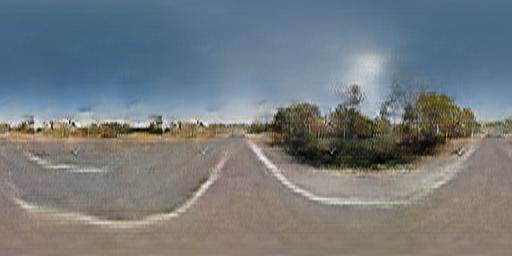} &
        \includegraphics[width=0.125\linewidth]{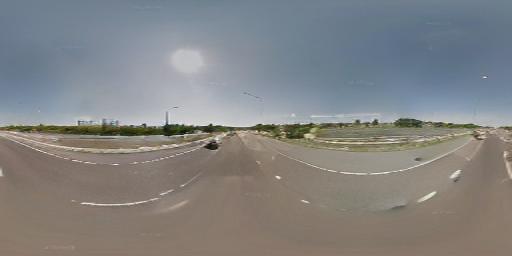} &
        \includegraphics[width=0.125\linewidth]{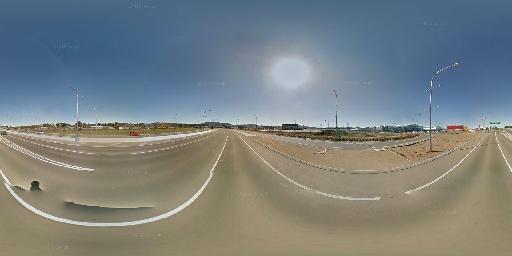}
        \\

    \end{tabular}
        \vspace{-0.2cm}
    \caption{Qualitative comparison of generated ground-level images on the CVACT dataset. We compare our method with ControlNet (CntrlNet), InstructPix2Pix (Inst P2P), BBDM, and Sat2Density (S2D).
    }
        \vspace{-0.3cm}
    \label{fig:cvact_visual}
\end{figure*}

\noindent{\textbf{CVACT.}} \figref{fig:cvact_visual} shows the contextual understanding and detail preservation of \name on the CVACT dataset. Our results demonstrate that incorporating explicit height maps into the conditioning process allows the synthesis of ground-level images with accurate height distributions and scene geometry. Specifically, all rows show that our model generates images with precise structural layouts, reflecting a precise reconstruction of height variations across different regions. The third and fourth rows reveal our model's ability to leverage aerial context. In the first row, the detailed reconstruction of a building contrasts with other baselines that tend to default to generic road scenes. This qualitative evidence confirms the robustness of our dual conditioning approach in producing context-aware, realistic ground-level images.

\begin{figure*}[ht]
    \centering
    \small
    \setlength{\tabcolsep}{0.5pt} %
    \renewcommand{\arraystretch}{0.1} %
    \begin{tabular}{cccccccc} %
        \multicolumn{1}{c}{Input} &
        \multicolumn{1}{c}{Height Map} &
        \multicolumn{1}{c}{CntrlNet} &
        \multicolumn{1}{c}{Inst P2P} &
        \multicolumn{1}{c}{BBDM} &
        \multicolumn{1}{c}{S2D} &
        \multicolumn{1}{c}{Ours} &
        \multicolumn{1}{c}{GT} \\

        \includegraphics[width=0.125\linewidth]{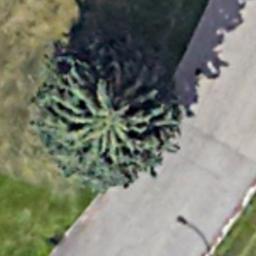} &
        \includegraphics[width=0.125\linewidth]{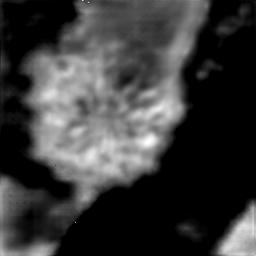} &
        \includegraphics[width=0.125\linewidth]{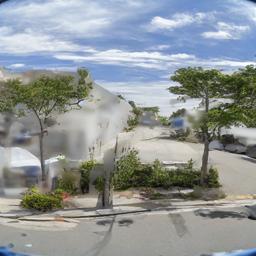} &
        \includegraphics[width=0.125\linewidth]{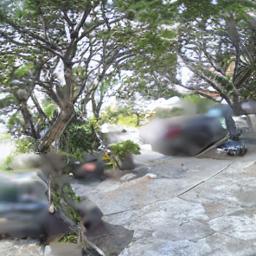} &
        \includegraphics[width=0.125\linewidth]{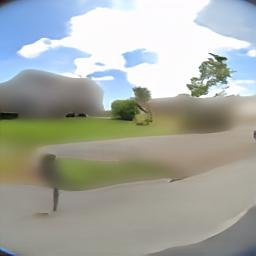} &
        \includegraphics[width=0.125\linewidth]{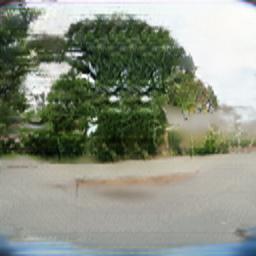} &
        \includegraphics[width=0.125\linewidth]{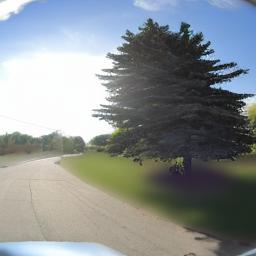} &
        \includegraphics[width=0.125\linewidth]{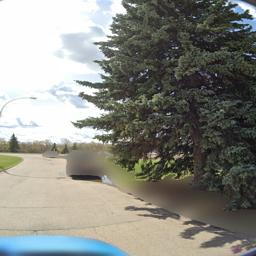}
        \\

        \includegraphics[width=0.125\linewidth]{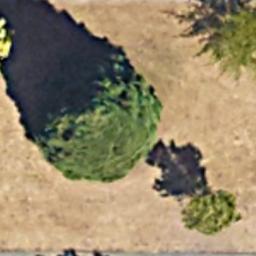} &
        \includegraphics[width=0.125\linewidth]{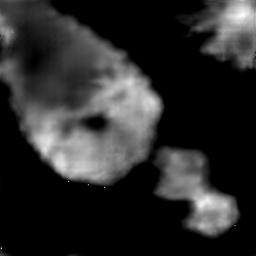} &
        \includegraphics[width=0.125\linewidth]{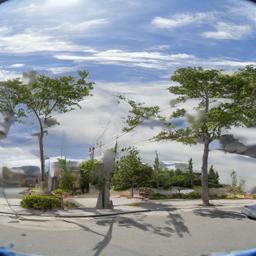} &
        \includegraphics[width=0.125\linewidth]{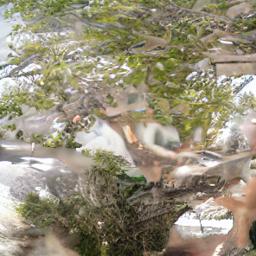} &
        \includegraphics[width=0.125\linewidth]{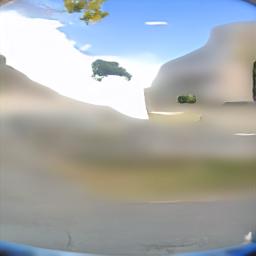} &
        \includegraphics[width=0.125\linewidth]{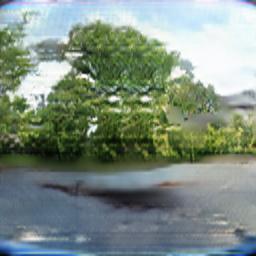} &
        \includegraphics[width=0.125\linewidth]{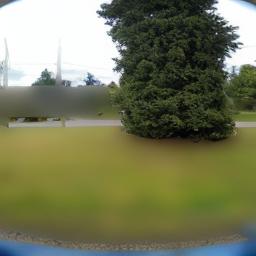} &
        \includegraphics[width=0.125\linewidth]{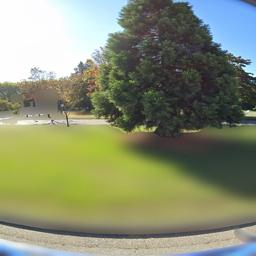}
        \\

        \includegraphics[width=0.125\linewidth]{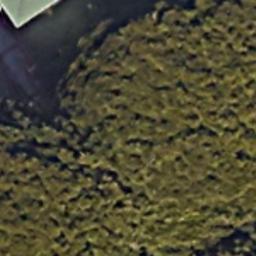} &
        \includegraphics[width=0.125\linewidth]{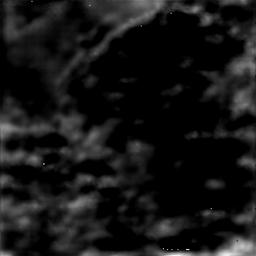} &
        \includegraphics[width=0.125\linewidth]{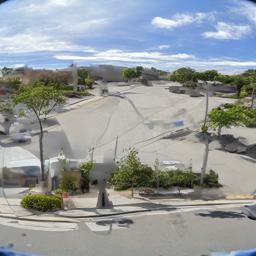} &
        \includegraphics[width=0.125\linewidth]{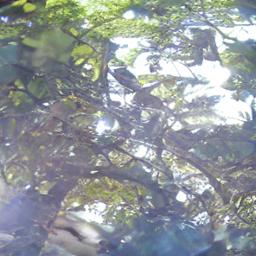} &
        \includegraphics[width=0.125\linewidth]{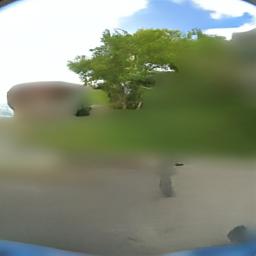} &
        \includegraphics[width=0.125\linewidth]{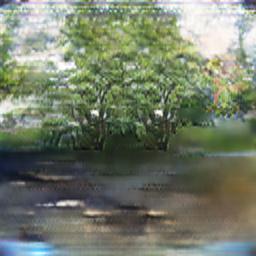} &
        \includegraphics[width=0.125\linewidth]{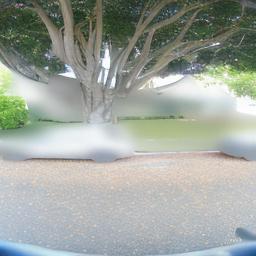} &
        \includegraphics[width=0.125\linewidth]{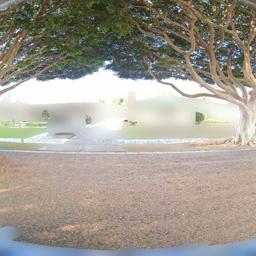}
        \\

        \includegraphics[width=0.125\linewidth]{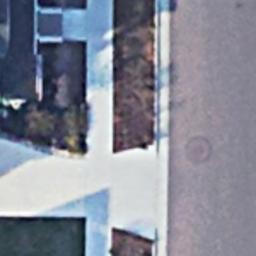} &
        \includegraphics[width=0.125\linewidth]{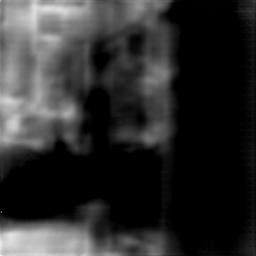} &
        \includegraphics[width=0.125\linewidth]{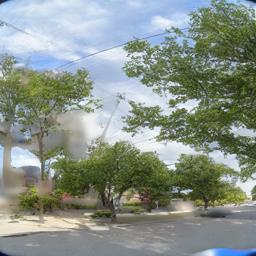} &
        \includegraphics[width=0.125\linewidth]{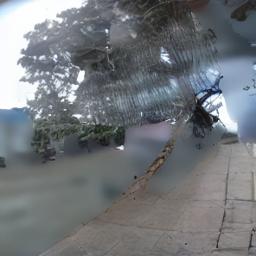} &
        \includegraphics[width=0.125\linewidth]{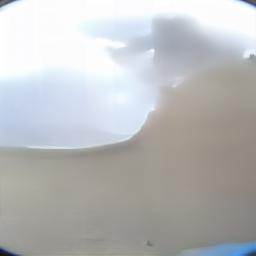} &
        \includegraphics[width=0.125\linewidth]{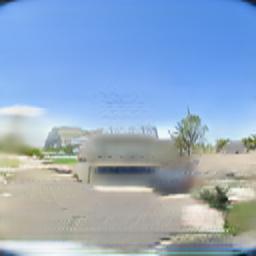} &
        \includegraphics[width=0.125\linewidth]{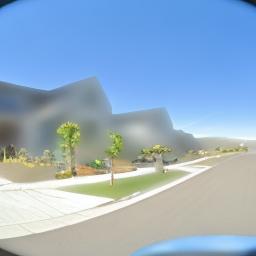} &
        \includegraphics[width=0.125\linewidth]{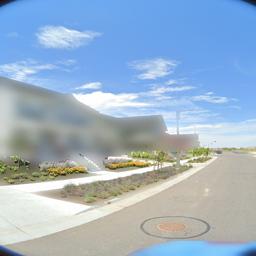}
        \\
    \end{tabular}
        \vspace{-0.3cm}
    \caption{
Qualitative comparison of generated ground-level images on the Auto Arborist dataset. We compare our method with ControlNet (CntrlNet), InstructPix2Pix (Inst P2P), BBDM, and Sat2Density (S2D).
    }
    \vspace{-0.4cm}
    \label{fig:aad_visual}
\end{figure*}

\noindent{\textbf{Auto Arborist Dataset.}} It is a more challenging scenario compared to the panoramic views of CVUSA~\cite{cvusa} and CVACT~\cite{cvact} due to its limited field of view from the top. Moreover, this constraint requires high precision in capturing scene details and object structures. As shown in~\Cref{fig:aad_visual}, our method consistently generates ground-level images that are both visually plausible and exact to the aerial input. For instance, in the fourth row, our model accurately reconstructs the tree branch style to match the ground truth, even though the input aerial image only reveals the tree canopy. Additionally, the first, second, and fourth rows demonstrate that our approach preserves fine-grained details, such as the texture of leaves and subtle variations in tree structure, and achieves higher image fidelity than baseline methods.

\subsection{Ablation Study}
\begin{table}[ht]
    \setlength{\tabcolsep}{2pt}
    \centering
    \resizebox{\linewidth}{!}{%
    \begin{NiceTabular}[color-inside]{cc|cccccc}
    \specialrule{.15em}{.05em}{.05em}
    CLIP & VAE & SSIM($\uparrow$) & IS($\uparrow$) & KID($\downarrow$) & Q-Align($\uparrow$) & CLIP($\uparrow$) & LPIPS($\downarrow$)\\ %
    \midrule
    \rowcolor{lightgray}
    \ding{51} & \ding{51}  & \textbf{0.50} & 2.63 & \textbf{0.06} & \textbf{2.12} & 0.75 & \textbf{0.55}\\
    \ding{51} & \ding{53}  & 0.34         & \textbf{2.64} & 0.07 & 2.05 & \textbf{0.76} & 0.63\\
    \ding{53} & \ding{51}  & 0.43         & 1.63          & 0.41 & 1.92 & 0.54 & 0.67\\
    \ding{53} & \ding{53}  & 0.34         & 1.59          & 0.16 & 2.02 & 0.49 & 0.66\\
    \specialrule{.15em}{.05em}{.05em}
    \end{NiceTabular}
    }
    \vspace{-0.3cm}
    \caption{An effect of inserting conditioning embedding spaces into our model using CVUSA~\cite{cvusa}. We put the \textbf{bold} to the best metrics.}
    \vspace{-0.3cm}
    \label{tab:ablation_condition}
\end{table}

We conduct three ablation studies on the CVUSA~\cite{cvusa} dataset to assess the contributions of each component in our architecture, including dual conditioning modules, classifier-free guidance scale, and the use of height map conditioning.

\textbf{Dual conditioning.}
\Cref{tab:ablation_condition} shows the combinations of the CLIP and VAE conditioning modules. Using both embeddings yields the best performance in six out of eight metrics, notably improving SSIM, KID, and LPIPS. These results validate the complementary roles of the VAE (for spatial detail) and CLIP (for semantic consistency) in achieving high-fidelity ground-view synthesis.

\begin{table}[hbt]
    \setlength{\tabcolsep}{2pt}
    \centering
    \resizebox{\linewidth}{!}{%
    \begin{NiceTabular}[color-inside]{c|cccccc}
    \specialrule{.15em}{.05em}{.05em}
    Guidance Scale & SSIM($\uparrow$) & IS($\uparrow$) & KID($\downarrow$) & Q-Align($\uparrow$) & CLIP($\uparrow$) & LPIPS($\downarrow$)\\ %
    \midrule
    1  & 0.42 & 2.57 & 0.07 & 2.05 & 0.76 & 0.59\\
    \rowcolor{lightgray}
    2  & \textbf{0.50} & 2.63 & \textbf{0.06} & 2.12 & 0.75 & \textbf{0.55}\\
    4  & 0.44 & 2.99 & 0.07 & 2.18 & 0.81 & 0.59\\
    8  & 0.43 & \textbf{3.30} & 0.07 & \textbf{2.19} & \textbf{0.82} & 0.60\\
    \specialrule{.15em}{.05em}{.05em}
    \end{NiceTabular}
    }
    \vspace{-0.1cm}
    \caption{An effect of guidance scale into our model using CVUSA~\cite{cvusa}. We put the \textbf{bold} to the best metrics.}
    \vspace{-0.3cm}
    \label{tab:ablation_cfg}
\end{table}

\textbf{Classifier-free guidance scale.}
As shown in \Cref{tab:ablation_cfg}, we vary the classifier-free guidance (CFG) scale across values \{1, 2, 4, 8\}. A guidance scale of 2 consistently achieves the best trade-off between structural integrity and perceptual quality, with peak SSIM and lowest LPIPS. Larger scales (e.g., 8) produce sharper images but at the cost of realism and stability, while smaller scales lead to blurrier results. These results highlight the importance of tuning CFG to optimize visual fidelity.

\begin{table}[hbt]
    \setlength{\tabcolsep}{2pt}
    \centering
    \resizebox{\linewidth}{!}{%
    \begin{NiceTabular}[color-inside]{c|cccccc}
    \specialrule{.15em}{.05em}{.05em}
    Height Map & SSIM($\uparrow$) & IS($\uparrow$) & KID($\downarrow$) & Q-Align($\uparrow$) & CLIP($\uparrow$) & LPIPS($\downarrow$)\\ %
    \midrule
    \rowcolor{lightgray}
    \ding{51}  & \textbf{0.50} & \textbf{2.63} & \textbf{0.06} & \textbf{2.12} & \textbf{0.75} & \textbf{0.55}\\
    \ding{53}  & 0.38          & 2.52          & 0.08          & 2.02          & 0.67          & 0.62\\
    \specialrule{.15em}{.05em}{.05em}
    \end{NiceTabular}
    }
    \vspace{-0.3cm}
    \caption{\small Without height map, performance drops significantly.}
    \vspace{-0.3cm}
    \label{tab:rebut_height_ab}
\end{table}

\begin{figure}[htb]
    \centering
    \small
    \scalebox{0.8}{ %
        \setlength{\tabcolsep}{0.5pt} %
        \renewcommand{\arraystretch}{0.1} %
        \begin{tabular}{cccc} %
            \multicolumn{1}{c}{Aerial} &
            \multicolumn{1}{c}{No Height} &
            \multicolumn{1}{c}{Yes Height} &
            \multicolumn{1}{c}{GT} \\
            \includegraphics[width=0.29\linewidth]{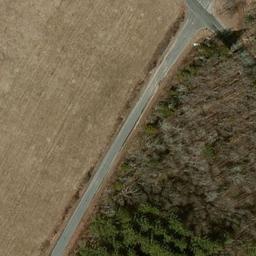} &
            \includegraphics[width=0.29\linewidth]{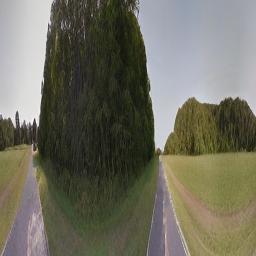} &
            \includegraphics[width=0.29\linewidth]{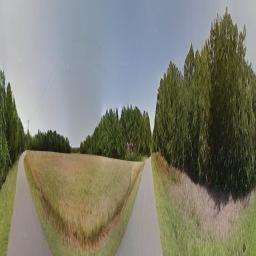} &
            \includegraphics[width=0.29\linewidth]{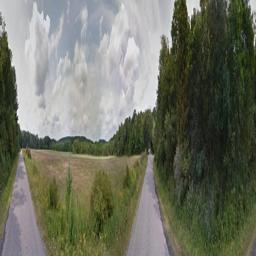}
            \\
        \end{tabular}
    } %
    \vspace{-0.2cm}
    \caption{Effect of removing height map conditioning. Without the height map, the model produces distorted ground-view images with degraded structural fidelity and incorrect object placements, highlighting the importance of spatial context for accurate synthesis.}
    \vspace{-0.3cm}
    \label{fig:rebut_no_height}
\end{figure}

\textbf{Height map conditioning.}
Removing the height map significantly degrades performance across all metrics (\Cref{tab:rebut_height_ab}), and qualitative examples (\Cref{fig:rebut_no_height}) show distorted object structures and incorrect layout. 
As shown in \cref{fig:rebut_no_height}, the model without the height prior produces a plausible-looking scene, but it fails because it does not match the geometric structure of the specific input aerial view. This shows the height prior's critical role as a geometric constraint, forcing the model to generate the correct scene rather than any plausible scene.
This underscores the critical role of spatial context provided by the height map in guiding accurate ground-view synthesis.

\textbf{Sensitivity to Height Map Noise.} We analyzed the model's sensitivity by adding Gaussian noise to the height maps on the CVUSA test set. Performance degraded, with SSIM dropping from 0.50 to 0.37. This result highlights the model's dependence on a quality geometric prior.

Together, these ablations confirm that each component, which is dual conditioning, CFG tuning, and height-awareness, is essential to the robustness and quality of \name. We will release the source code and the model upon acceptance.
\begin{figure}[htb]
    \centering
    \small
    \setlength{\tabcolsep}{0.5pt} %
    \renewcommand{\arraystretch}{0.1} %
    \begin{tabular}{cccccccc} %
        \multicolumn{1}{c}{Input} &
        \multicolumn{1}{c}{Height} &
        \multicolumn{1}{c}{CntrlNet} &
        \multicolumn{1}{c}{Inst P2P} &
        \multicolumn{1}{c}{BBDM} &
        \multicolumn{1}{c}{S2D} &
        \multicolumn{1}{c}{Ours} &
        \multicolumn{1}{c}{GT} \\

        \includegraphics[width=0.125\linewidth]{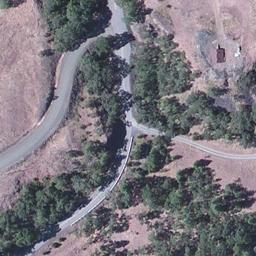} &
        \includegraphics[width=0.125\linewidth]{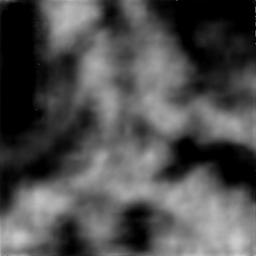} &
        \includegraphics[width=0.125\linewidth]{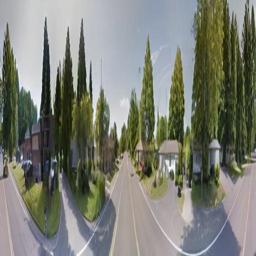} &
        \includegraphics[width=0.125\linewidth]{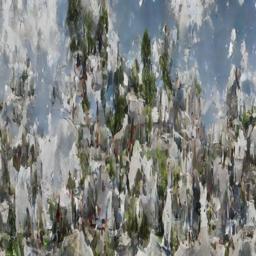} &
        \includegraphics[width=0.125\linewidth]{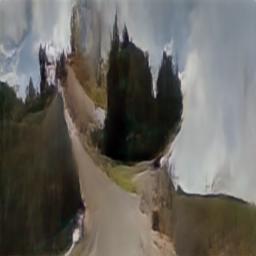} &
        \includegraphics[width=0.125\linewidth]{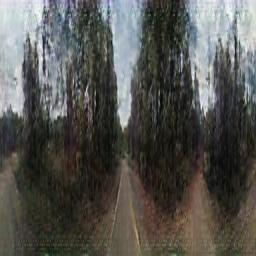} &
        \includegraphics[width=0.125\linewidth]{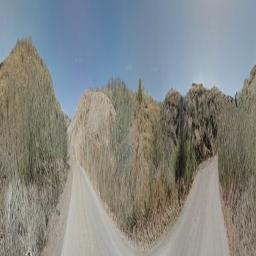} &
        \includegraphics[width=0.125\linewidth]{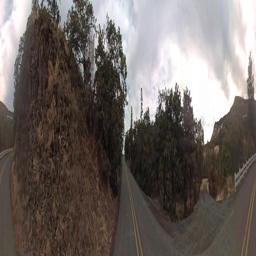}
        \\

        \includegraphics[width=0.125\linewidth]{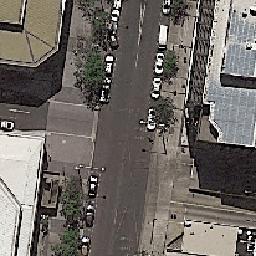} &
        \includegraphics[width=0.125\linewidth]{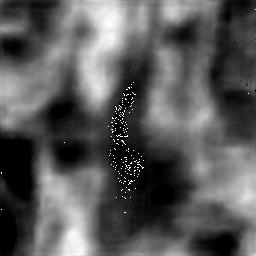} &
        \includegraphics[width=0.125\linewidth]{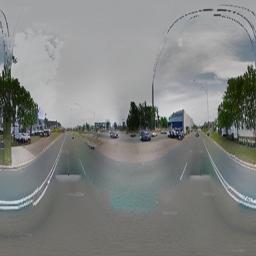} &
        \includegraphics[width=0.125\linewidth]{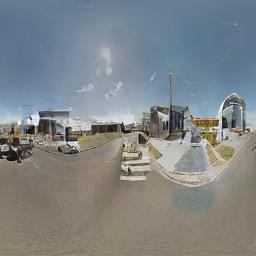} &
        \includegraphics[width=0.125\linewidth]{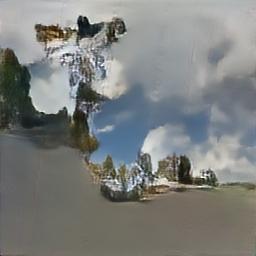} &
        \includegraphics[width=0.125\linewidth]{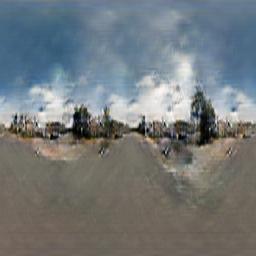} &
        \includegraphics[width=0.125\linewidth]{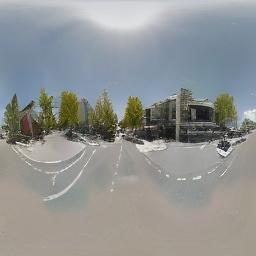} &
        \includegraphics[width=0.125\linewidth]{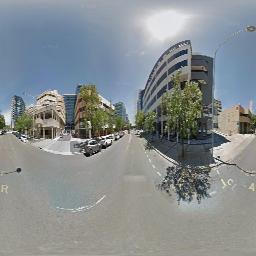}
        \\

        \includegraphics[width=0.125\linewidth]{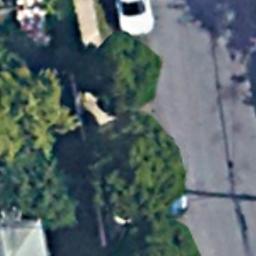} &
        \includegraphics[width=0.125\linewidth]{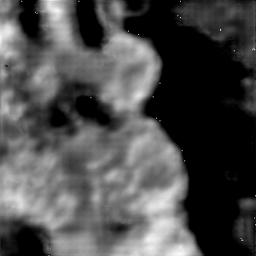} &
        \includegraphics[width=0.125\linewidth]{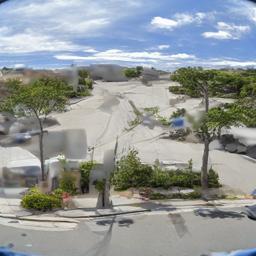} &
        \includegraphics[width=0.125\linewidth]{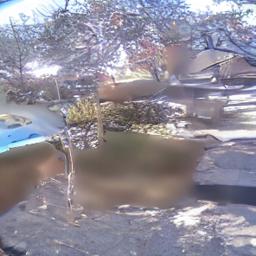} &
        \includegraphics[width=0.125\linewidth]{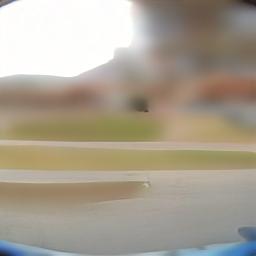} &
        \includegraphics[width=0.125\linewidth]{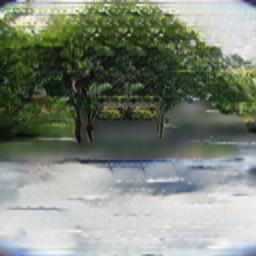} &
        \includegraphics[width=0.125\linewidth]{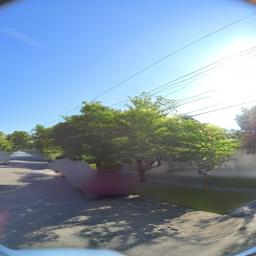} &
        \includegraphics[width=0.125\linewidth]{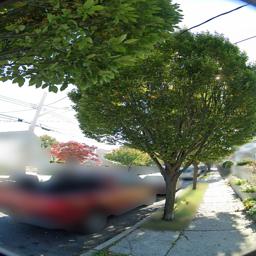}
        \\

    \end{tabular}
    \vspace{-0.3cm}
    \caption{Failure cases from CVUSA (1st row), CVACT (2nd row), and AAD (3rd row) from ours and ControlNet (CntrlNet), Instruct Pix2Pix (Inst P2P), BBDM, Sat2Density (S2D).
    }
    \vspace{-0.4cm}
    \label{fig:failures}
\end{figure}

\subsection{Limitation and Future Work}
\name\ relies on an aerial RGB image~$x$ and its corresponding estimated height map $H(x)$ from a pre-trained model. Thus, the quality of the generated ground-level image $y$ is naturally connected to the accuracy of $H(x)$. When height estimation is suboptimal, our model's output may show degraded structural fidelity and inaccurate object representations. As we show our failure cases in~\Cref{fig:failures}, the first row (CVUSA~\cite{cvusa}) shows that ours get similar height contributions on our image, but the leaf details are off. The second row (CVACT~\cite{cvact}) shows that our work generates buildings, but their height is not tall enough. The last row (AAD~\cite{aad}) shows that it gets a tree instance, and its placement is off.

As future work, we plan to extend \name\ by combining object-specific priors, which help to improve image fidelity by allowing the model to capture fine-grained details and complicated scene semantics better. Moreover, leveraging different sensor modalities, such as thermal or hyperspectral data, could improve the robustness and quality of ground-level generation under challenging conditions by providing extra information. These additional modalities can broaden the applicability of our model across diverse and dynamic real-world environments.

\section{Conclusion}
We introduced \name, a novel diffusion-based framework for aerial-to-ground view synthesis that leverages height-aware dual conditioning. By integrating VAE-based spatial features and CLIP-based semantic cues via cross-attention, our model synthesizes high-fidelity ground-level images directly from aerial inputs, without relying on intermediate representations such as 3D voxels or density maps. This design enables both structural accuracy and semantic consistency while maintaining computational efficiency. Moreover, the modular conditioning framework supports extensibility to other modalities (e.g., thermal, multispectral, or object-level priors), facilitating broader applications in remote sensing and environmental modeling.

Extensive evaluations on CVUSA~\cite{cvusa}, CVACT~\cite{cvact}, and the Auto Arborist Dataset~\cite{aad} demonstrate consistent improvements over state-of-the-art methods across perceptual, structural, and pixel-level metrics. In particular, performance gains on AAD highlight \name's robustness in narrow field-of-view, underscoring its practicality for real-world UAV deployment.

Future work could incorporate object-aware priors and temporal consistency to enhance synthesis realism and fine-grained control further. \name could offer a scalable and effective foundation for future advances in cross-view generation, urban simulation, and geo-spatial AI.

\clearpage
{
    \small
    \bibliographystyle{ieeenat_fullname}
    \bibliography{main}

@String(PAMI = {IEEE Trans. Pattern Anal. Mach. Intell.})

@String(CVPR= {IEEE Conf. Comput. Vis. Pattern Recog.})

@String(ICCV= {Int. Conf. Comput. Vis.})

@String(ECCV= {Eur. Conf. Comput. Vis.})

@String(NIPS= {Adv. Neural Inform. Process. Syst.})

@String(ICLR = {Int. Conf. Learn. Represent.})

@String(AAAI = {AAAI})

@String(ICML = {ICML})

@String(PAMI  = {IEEE TPAMI})

@String(CVPR  = {CVPR})

@String(ICCV  = {ICCV})

@String(ECCV  = {ECCV})

@String(NIPS  = {NeurIPS})

@String(ICLR  = {ICLR})

@inproceedings{BBDM,
  title={Bbdm: Image-to-image translation with brownian bridge diffusion models},
  author={Li, Bo and Xue, Kaitao and Liu, Bin and Lai, Yu-Kun},
  booktitle=CVPR,
  year={2023}
}

@inproceedings{controlnet,
  title={Adding conditional control to text-to-image diffusion models},
  author={Zhang, Lvmin and Rao, Anyi and Agrawala, Maneesh},
  booktitle=ICCV,
  year={2023}
}

@inproceedings{instructpix2pix,
  title={Instructpix2pix: Learning to follow image editing instructions},
  author={Brooks, Tim and Holynski, Aleksander and Efros, Alexei A},
  booktitle=CVPR,
  year={2023}
}

@inproceedings{sat2density,
  title={Sat2density: Faithful density learning from satellite-ground image pairs},
  author={Qian, Ming and Xiong, Jincheng and Xia, Gui-Song and Xue, Nan},
  booktitle=CVPR,
  year={2023}
}

@inproceedings{treedfusion,
  title={Tree-D Fusion: Simulation-Ready Tree Dataset from Single Images with Diffusion Priors},
  author={Lee, Jae Joong and Li, Bosheng and Beery, Sara and Huang, Jonathan and Fei, Songlin and Yeh, Raymond A and Benes, Bedrich},
  booktitle=ECCV,
  year={2025},
}

@inproceedings{cvusa,
  title={Wide-area image geolocalization with aerial reference imagery},
  author={Workman, Scott and Souvenir, Richard and Jacobs, Nathan},
  booktitle=ICCV,
  year={2015}
}

@inproceedings{cvact,
  title={Optimal feature transport for cross-view image geo-localization},
  author={Shi, Yujiao and Yu, Xin and Liu, Liu and Zhang, Tong and Li, Hongdong},
  booktitle=AAAI,
  year={2020}
}

@inproceedings{aad,
  title={The auto arborist dataset: a large-scale benchmark for multiview urban forest monitoring under domain shift},
  author={Beery, Sara and Wu, Guanhang and Edwards, Trevor and Pavetic, Filip and Majewski, Bo and Mukherjee, Shreyasee and Chan, Stanley and Morgan, John and Rathod, Vivek and Huang, Jonathan},
  booktitle=CVPR,
  year={2022}
}

@article{QAlign,
  title={Q-align: Teaching lmms for visual scoring via discrete text-defined levels},
  author={Wu, Haoning and Zhang, Zicheng and Zhang, Weixia and Chen, Chaofeng and Liao, Liang and Li, Chunyi and Gao, Yixuan and Wang, Annan and Zhang, Erli and Sun, Wenxiu and others},
  journal=ICML,
  year={2024}
}

@inproceedings{crossviewdiff,
author = {Chen, Yuankun and Rong, Dazhong and Li, Yi},
title = {CrossViewDiff: A Cross-View Diffusion Model for Satellite-to-Ground Image Synthesis},
year = {2024},
booktitle = {International Conference on Artificial Neural Networks},
}

@article{nichol2021glide,
  title={Glide: Towards photorealistic image generation and editing with text-guided diffusion models},
  author={Nichol, Alex and Dhariwal, Prafulla and Ramesh, Aditya and Shyam, Pranav and Mishkin, Pamela and McGrew, Bob and Sutskever, Ilya and Chen, Mark},
  journal={arXiv preprint arXiv:2112.10741},
  year={2021}
}

@inproceedings{liu2023zero,
  title={Zero-1-to-3: Zero-shot one image to 3d object},
  author={Liu, Ruoshi and Wu, Rundi and Van Hoorick, Basile and Tokmakov, Pavel and Zakharov, Sergey and Vondrick, Carl},
  booktitle=ICCV,
  year={2023}
}

@article{watson2022novel,
  title={Novel view synthesis with diffusion models},
  author={Watson, Daniel and Chan, William and Martin-Brualla, Ricardo and Ho, Jonathan and Tagliasacchi, Andrea and Norouzi, Mohammad},
  journal={arXiv preprint arXiv:2210.04628},
  year={2022}
}

@article{cfg,
  title={Classifier-free diffusion guidance},
  author={Ho, Jonathan and Salimans, Tim},
  journal={arXiv preprint arXiv:2207.12598},
  year={2022}
}

@article{poole2022dreamfusion,
  title={Dreamfusion: Text-to-3d using 2d diffusion},
  author={Poole, Ben and Jain, Ajay and Barron, Jonathan T and Mildenhall, Ben},
  journal={arXiv preprint arXiv:2209.14988},
  year={2022}
}

@inproceedings{rombach2022high,
  title={High-resolution image synthesis with latent diffusion models},
  author={Rombach, Robin and Blattmann, Andreas and Lorenz, Dominik and Esser, Patrick and Ommer, Bj{\"o}rn},
  booktitle=CVPR,
  year={2022}
}

@inproceedings{clip,
  title={Learning transferable visual models from natural language supervision},
  author={Radford, Alec and Kim, Jong Wook and Hallacy, Chris and Ramesh, Aditya and Goh, Gabriel and Agarwal, Sandhini and Sastry, Girish and Askell, Amanda and Mishkin, Pamela and Clark, Jack and others},
  booktitle=ICML,
  year={2021},
}

@article{shi2022geometry,
  title={Geometry-guided street-view panorama synthesis from satellite imagery},
  author={Shi, Yujiao and Campbell, Dylan and Yu, Xin and Li, Hongdong},
  journal=PAMI,
  year={2022},

}

@inproceedings{zhai2017predicting,
  title={Predicting ground-level scene layout from aerial imagery},
  author={Zhai, Menghua and Bessinger, Zachary and Workman, Scott and Jacobs, Nathan},
  booktitle=CVPR,
  year={2017}
}

@inproceedings{tang2019multi,
  title={Multi-channel attention selection gan with cascaded semantic guidance for cross-view image translation},
  author={Tang, Hao and Xu, Dan and Sebe, Nicu and Wang, Yanzhi and Corso, Jason J and Yan, Yan},
  booktitle=CVPR,
  year={2019}
}

@inproceedings{li2021sat2vid,
  title={Sat2vid: Street-view panoramic video synthesis from a single satellite image},
  author={Li, Zuoyue and Li, Zhenqiang and Cui, Zhaopeng and Qin, Rongjun and Pollefeys, Marc and Oswald, Martin R},
  booktitle=CVPR,
  year={2021}
}

@inproceedings{regmi2018cross,
  title={Cross-view image synthesis using conditional gans},
  author={Regmi, Krishna and Borji, Ali},
  booktitle=CVPR,
  year={2018}
}

@inproceedings{lu2020geometry,
  title={Geometry-aware satellite-to-ground image synthesis for urban areas},
  author={Lu, Xiaohu and Li, Zuoyue and Cui, Zhaopeng and Oswald, Martin R and Pollefeys, Marc and Qin, Rongjun},
  booktitle=CVPR,
  year={2020}
}

@inproceedings{cvact_original,
  title={Lending orientation to neural networks for cross-view geo-localization},
  author={Liu, Liu and Li, Hongdong},
  booktitle=CVPR,
  year={2019}
}

@inproceedings{zhu2021vigor,
  title={Vigor: Cross-view image geo-localization beyond one-to-one retrieval},
  author={Zhu, Sijie and Yang, Taojiannan and Chen, Chen},
  booktitle=CVPR,
  year={2021}
}

@article{ho2020denoising,
  title={Denoising diffusion probabilistic models},
  author={Ho, Jonathan and Jain, Ajay and Abbeel, Pieter},
  journal=NIPS,
  year={2020}
}

@inproceedings{
  song2021scorebased,
  title={Score-Based Generative Modeling through Stochastic Differential Equations},
  author={Yang Song and Jascha Sohl-Dickstein and Diederik P Kingma and Abhishek Kumar and Stefano Ermon and Ben Poole},
  booktitle=ICLR,
  year={2021},
}

@article{dhariwal2021diffusion,
  title={Diffusion models beat gans on image synthesis},
  author={Dhariwal, Prafulla and Nichol, Alexander},
  journal=NIPS,
  year={2021}
}

@inproceedings{kim2022diffusionclip,
  title={Diffusionclip: Text-guided diffusion models for robust image manipulation},
  author={Kim, Gwanghyun and Kwon, Taesung and Ye, Jong Chul},
  booktitle=CVPR,
  year={2022}
}

@article{saharia2022photorealistic,
  title={Photorealistic text-to-image diffusion models with deep language understanding},
  author={Saharia, Chitwan and Chan, William and Saxena, Saurabh and Li, Lala and Whang, Jay and Denton, Emily L and Ghasemipour, Kamyar and Gontijo Lopes, Raphael and Karagol Ayan, Burcu and Salimans, Tim and others},
  journal=NIPS,
  year={2022}
}

@article{cambrin2024depth,
  title={Depth any canopy: Leveraging depth foundation models for canopy height estimation},
  author={Cambrin, Daniele Rege and Corley, Isaac and Garza, Paolo},
  journal={arXiv preprint arXiv:2408.04523},
  year={2024}
}

@misc{faa_drones,
  author       = {Federal Aviation Administration},
  title        = {Drones by the Numbers},
  year         = {2025},
  note          = {\url{https://www.faa.gov/node/54496}},
}

@misc{us_drone_growth,
  author       = {European Agency for Safety and Health at Work},
  title        = {UNMANNED AERIAL VEHICLES: IMPLICATIONS FOR
OCCUPATIONAL SAFETY AND HEALTH},
  year         = {2022},
  note          = {\url{https://osha.europa.eu/sites/default/files/Unnamed-aerial-vehicles-and-OSH_en.pdf}},
}

@article{loshchilov2017decoupled,
  title={Decoupled weight decay regularization},
  author={Loshchilov, Ilya and Hutter, Frank},
  journal=ICLR,
  year={2019}
}

@inbook{RegeCambrin2025,
  title = {Depth Any Canopy: Leveraging Depth Foundation Models for Canopy Height Estimation},
  ISBN = {9783031923876},
  ISSN = {1611-3349},
  url = {http://dx.doi.org/10.1007/978-3-031-92387-6_5},
  DOI = {10.1007/978-3-031-92387-6_5},
  booktitle = {Computer Vision – ECCV 2024 Workshops},
  author = {Rege Cambrin,  Daniele and Corley,  Isaac and Garza,  Paolo},
  year = {2025},
}

@inproceedings{velazquez2025earthview,
  title={EarthView: a large scale remote sensing dataset for self-supervision},
  author={Velazquez, Diego and Rodriguez, Pau and Alonso, Sergio and Gonfaus, Josep M and Gonzalez, Jordi and Richarte, Gerardo and Marin, Javier and Bengio, Yoshua and Lacoste, Alexandre},
  booktitle={Proceedings of the Winter Conference on Applications of Computer Vision},
  year={2025}
}

@inproceedings{liu2024grounding,
  title={Grounding dino: Marrying dino with grounded pre-training for open-set object detection},
  author={Liu, Shilong and Zeng, Zhaoyang and Ren, Tianhe and Li, Feng and Zhang, Hao and Yang, Jie and Jiang, Qing and Li, Chunyuan and Yang, Jianwei and Su, Hang and others},
  booktitle=ECCV,
  year={2024},
}
}
\clearpage


\end{document}